\documentclass{article}
\pdfoutput=1
\usepackage{url}            
\usepackage{booktabs}       
\usepackage{amsfonts}       
\usepackage{nicefrac}       
\usepackage[utf8]{inputenc}
\usepackage{multirow}
\usepackage{microtype}      
\usepackage{xcolor}         
\usepackage[preprint]{neurips_2025_data}
\usepackage{amsmath}
\usepackage{graphicx}
\usepackage{authblk}
\usepackage{subcaption}
\usepackage{colortbl} 
\usepackage{hhline}
\usepackage{fontawesome}
\usepackage{caption}
\usepackage{subcaption}
\usepackage{placeins}
\PassOptionsToPackage{numbers,sort&compress}{natbib}
\usepackage[numbers,sort&compress]{natbib}
\usepackage{wrapfig}

\usepackage{authblk}                 
\usepackage{hyperref}     
\usepackage{listings}
\newcommand{\benchmarkname}{{HSE-Bench}}

\makeatletter
  \let\oldthanks\thanks
  \renewcommand\thanks[1]{%
    \oldthanks{#1}\kern0.4em
  }
\makeatother

\title{LLM-based HSE Compliance Assessment: Benchmark, Performance, and Advancements}

\hypersetup{
    colorlinks=true,
    urlcolor=magenta,  
}

\author[1]{Jianwei Wang\thanks{jianwei.wang1@unsw.edu.au}}
\author[1]{Mengqi Wang}
\author[1]{Yinsi Zhou}
\author[2]{Zhenchang Xing}
\author[2]{Qing Liu}
\author[2]{Xiwei Xu}
\author[1]{ Wenjie Zhang}
\author[2]{Liming Zhu}

\affil[1]{University of New South Wales}
\affil[2]{Data61, CSIRO}
\affil[ ]{\url{https://huggingface.co/datasets/Joysouo/hse-bench}}
\affil[ ]{\url{https://github.com/mengqiwang1/hse-bench}}

\begin{document}

\maketitle
\vspace{-2mm}
\begin{abstract}
Health, Safety, and Environment (HSE) compliance assessment demands dynamic real-time decision-making under complicated regulations and complex human-machine-environment interactions. While large language models (LLMs) hold significant potential for decision intelligence and contextual dialogue, their capacity for domain-specific knowledge in HSE and structured legal reasoning remains underexplored. 
We introduce HSE-Bench, the first benchmark dataset designed to evaluate the HSE compliance assessment capabilities of LLM. 
HSE-Bench comprises over 1,000 manually curated questions drawn from regulations, court cases, safety exams, and fieldwork videos, and integrates a reasoning flow based on Issue spotting, rule Recall, rule Application, and rule Conclusion (IRAC) to assess the holistic reasoning pipeline.
We conduct extensive evaluations on different prompting strategies and more than 10 LLMs, including foundation models, reasoning models and multimodal vision models.
The results show that, although current LLMs achieve good performance, their capabilities largely rely on semantic matching rather than principled reasoning grounded in the underlying HSE compliance context.
Moreover, their native reasoning trace lacks the systematic legal reasoning required for rigorous HSE compliance assessment.
To alleviate these, we propose a new prompting technique, Reasoning of Expert (RoE), which guides LLMs to simulate the reasoning process of different experts for compliance assessment and reach a more accurate unified decision.
We hope our study highlights reasoning gaps in LLMs for HSE compliance and inspires further research on related tasks.

\end{abstract}

\vspace{-3mm}
\section{Introduction}
\label{sec:intro}
\vspace{-3mm}
The rapid advancement of Large Language Models (LLMs) has sparked significant interest in assessing their performance across a wide range of real-world scenarios~\cite{guha2023legalbench, khandekar2024medcalc,liu2024synthetic, wang2025llm,zhang2023safetybench}. 
To support structured evaluations of LLM capabilities in various domains, several benchmarks have been proposed~\cite{pramanick2024spiqa}. For instance, LEGALBENCH~\cite{guha2023legalbench}, MEDCALC-BENCH~\cite{khandekar2024medcalc}, and BEACON~\cite{ren2024beacon} offer domain-specific assessments in law, medical science and biology, respectively.
These benchmarks play a critical role in evaluating the capabilities and limitations of LLMs, facilitating fair comparison, performance assessment, and progress tracking across different models.



The capability of HSE (Health, Safety and Environment) compliance assessment enables an organization to establish and refine management systems, processes and controls for managing health, safety and environmental risks~\cite{akamangwa2016working,arewa2012review,duijm2008management}. 
Failure to conduct this HSE compliance assessment may jeopardize employee well-being and lead to non-compliance with legal and regulatory obligations, posing significant risks for stakeholders in industries such as construction, manufacturing, and healthcare.
Although LLMs demonstrate significant potential for decision intelligence and contextual dialogue across various vertical domains, as evidenced by numerous benchmarks~\cite{abdelnabi2024cooperation, chang2024survey, shao2024visual, xie2024finben}, their capacity for HSE compliance assessment remains largely unexplored~\cite{shirkhorshidi2023comply}. Without a robust evaluation framework, using LLMs in HSE contexts may yield unreliable or non‑compliant recommendations, potentially leading to severe risks in high‑stakes environments.
On the other hand, building a benchmark for HSE compliance assessment is highly challenging, including 
(1) \textbf{diverse data sources}: in real-world HSE contexts, queries may originate from various sources such as video recordings of inspection sites and textual descriptions, which require a unified understanding of the data~\cite{hajipour2021integrated, xu2022smart}; 
(2) \textbf{complicated scenarios}: real-world HSE tasks often involve a broad range of high-risk and high-impact situations that require nuanced and context-aware analysis~\cite{hojageldiyev2018artificial}; 
(3) \textbf{complex reasoning processes}: effective HSE decision-making frequently relies on multi-step reasoning, such as issue spotting and relevant regulation recall, which significantly increases the difficulty of benchmark construction~\cite{shirkhorshidi2023comply,zhang2015ontology}.
To address these issues, a comprehensive HSE compliance assessment benchmark is necessary to evaluate LLMs rigorously, ensuring they perform reliably across various HSE contexts and contribute to safer, more effective decision-making in HSE. A recent work~\cite{ehs-llm} makes some initial explorations in this direction. However, it relies on limited data sources and scenarios and lacks a integrated reasoning process, which restricts its effectiveness and applicability.

Our research aims to fill this gap by proposing \benchmarkname, a new and systematic approach for benchmarking, performance evaluation, and advancing the application of LLMs in HSE compliance assessment. 
Firstly, we create a comprehensive, curated benchmark dataset for HSE reasoning by transforming diverse, authoritative, and multi-modal sources, including regulations, court cases, safety exams, and fieldwork videos, into structured, scenario-based assessments with labels. 
Our question generation strategy first follows the IRAC framework~\cite{choi2021chatgpt,emiri2017revisiting,guha2023legalbench}, structuring questions across four reasoning dimensions: \textbf{I}ssue spotting, rule \textbf{R}ecall, rule \textbf{A}pplication, and rule \textbf{C}onclusion. 
To enhance the quality of generated questions, we further introduce the adversarial generation strategy to include distractors w.r.t. the options, add logical complexity and remove confidential data.
Then, two experts manually check each question.
This approach enables comprehensive benchmarking of LLM performance on HSE-related questions involving diverse data sources and multi-step reasoning.
Secondly, we evaluate the performance of various state-of-the-art LLMs, covering both the leading foundation models, reasoning models and vision models. We also evaluate different promoting strategies, including Zero-shot, Few-shot~\cite{brown2020language}, Chain of Thought (CoT)~\cite{wei2022chain}, Zero CoT~\cite{wang2023plan} and CoT with Self-Consistency~\cite{wangself}.
The evaluation results reveal that (1) although current LLMs achieve strong performance, their capabilities primarily rely on semantic matching rather than principled reasoning grounded in the underlying HSE compliance context, and (2) their native reasoning trace lacks the systematic legal reasoning required for rigorous HSE compliance assessment.
Thirdly, we propose a new prompting technique, the Reasoning of Experts (RoE). It mirrors real-world HSE legal reasoning processes by having the model assume different roles of domain-specific experts, each providing independent reasoning judgments, which are then synthesized into a final decision.


In summary, the contributions of this paper are threefold:

\begin{itemize}
\item \textbf{Comprehensive HSE Benchmark:} We create a curated HSE reasoning benchmark with scenario-based assessments from diverse sources, using the IRAC framework and adversarial generation strategies to evaluate LLM performance.
\item \textbf{Evaluation of LLM Performance:} We evaluate leading LLMs using various prompting strategies, finding that current models rely on semantic matching rather than true legal understanding and they lack systematic HSE compliance reasoning.
\item \textbf{RoE Technique:} We introduce the RoE prompting, where models assume different roles of domain experts to synthesize independent reasoning into a final decision, improving the capacity of LLMs for HSE compliance.

\end{itemize}


\section{Related Works}
\label{sec:related_works}




\noindent\textbf{LLM Evaluations in Regulation Scenarios}. 
A set of works is proposed to evaluate LLMs in regulation-intensive domains, where factual accuracy, reasoning fidelity, and behavioral alignment are critical. 
These benchmarks often adopt structured and systematic taxonomies (e.g., cognitive layers and IRAC), revealing valuable insights. 
Benchmarks like LegalBench~\cite{guha2023legalbench}, LawBench~\cite{zheng2023lawbench}, and LAiW~\cite{zhang2024laiw} decompose legal tasks into fine-grained capabilities such as issue spotting and legal consultation, allowing for a comprehensive evaluation of model performance. A recent study~\cite{ehs-llm} explores fine-tuning LLMs for HSE applications.
Beyond legal scenarios, recent efforts have expanded the scope to more general regulation-centric contexts.
A financial regulation benchmark is proposed in~\cite{wang2024professional} to assess LLM outputs against complex regulatory information.
RuleArena~\cite{jiang2025rulearena} generalizes these efforts and evaluates three concrete rule- and regulation-centric scenarios: airline baggage fees, NBA transactions, and tax regulations. It focuses on rule-grounded reasoning over real-world policy texts, revealing that even advanced LLMs struggle to apply and compose natural language rules in multi-step reasoning chains.

Beyond evaluating the correctness of LLMs, recent studies have investigated their trustworthiness and behavioral dynamics. One study~\cite{dahl2024legalhallucinations} shows that LLMs systematically hallucinate legal facts, especially in low-resource judicial settings. FairBalancing~\cite{fairbalancing2024} explores ethical judgments in privacy policy analysis, revealing instabilities across prompt framings and value perspectives. RegSum~\cite{regsum2024} highlights trade-offs between fluency and factual faithfulness in document-level summarization of long-form regulatory texts.



\noindent\textbf{LLM Evaluations in High-risk Scenarios}. Risk-sensitive contexts are vital in real-world applications where mistakes can cause serious harm. LLMs are being explored to enhance decision-making in these contexts. 
In scientific domains, LabSafety Bench~\cite{zhou2024labsafety} is designed to identify potential hazards, assess risks, and predict the consequences of unsafe actions in laboratory settings. AIR-BENCH~\cite{zengair} evaluates whether LLMs comply with specific regulations and policies.
In the industrial context, the H\&PS Traffic Incidents Dataset~\cite{li2025llms} is introduced to evaluate the performance of LLMs in handling traffic incidents.
LLMs are also applied to automate accident analysis and safety diagnostics~\cite{aips2024}, enabling rapid incident classification and mitigation recommendations. They have also been employed in equipment failure detection~\cite{equipmentfailure2024}, construction site compliance checks~\cite{twostagerag2024}, and personal protective equipment (PPE) verification~\cite{clip2safety2024}.
Additionally, VR-augmented environments~\cite{vrhazards2024} and LLM-assisted safety learning systems~\cite{isafechatbot2024} explore new modalities for enhancing safety awareness and education, particularly through multi-turn, multi-source integration.



Despite these advances, existing benchmarks still have several limitations, particularly in handling diverse data sources, complex contexts, and tasks that require sophisticated legal reasoning. To address these, we propose \benchmarkname, a new and systematic framework for benchmarking, performance evaluation, and advancing the application of LLMs in HSE scenarios.

\vspace{-2mm}
\section{HSE-BENCH}
\vspace{-2mm}
\label{sec:method}

\begin{figure}
    \centering
    \includegraphics[width=\linewidth]{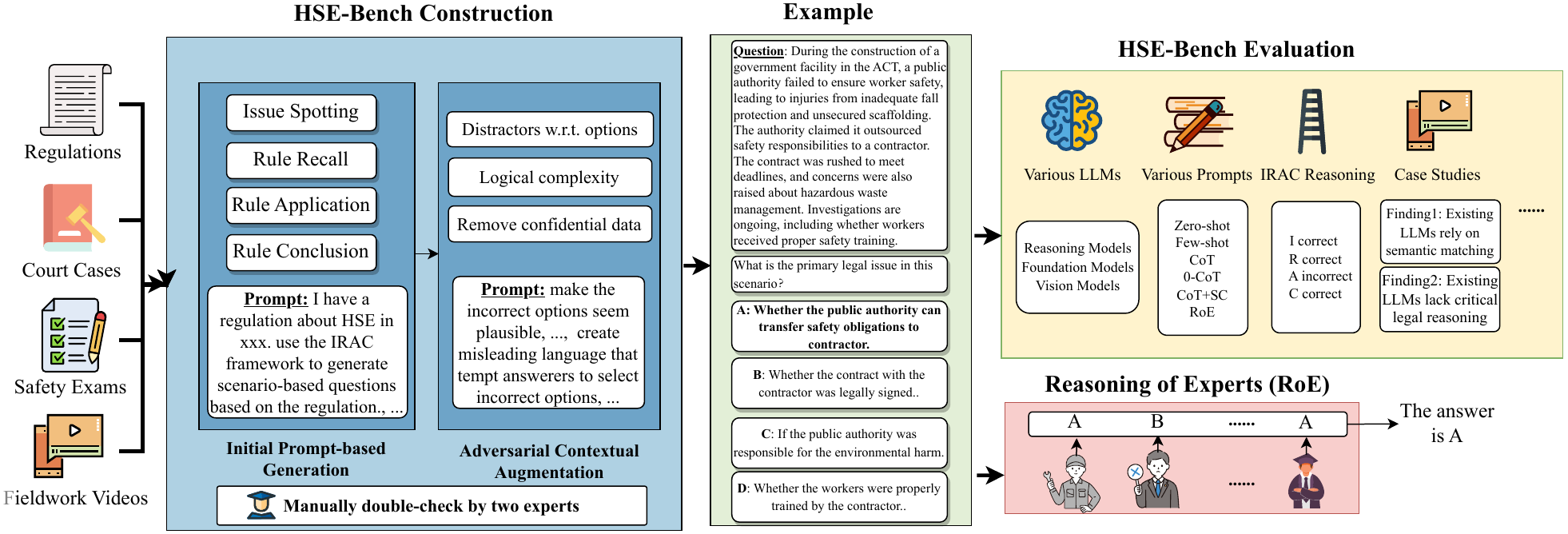}
    \caption{Conceptual framework of this project, featuring benchmark construction, comprehensive performance evaluation and new advancement to apply LLM for HSE compliance assessment}
    \vspace{-4mm}
    \label{fig:framework}
\end{figure}

The overall framework of \benchmarkname~is shown in Figure~\ref{fig:framework}. We construct dataset instances from four sources: regulations, court cases, safety exams, and fieldwork videos. Based on the IRAC taxonomy, we follow a two-phase benchmark construction process. First, we use GPT-4o to generate scenario-based questions from the extracted data. Then, we apply adversarial contextual augmentation to further improve question quality.
To evaluate performance, we test 12 LLMs, covering both advanced reasoning models and foundation models. We also explore various prompts and settings to ensure a thorough evaluation. To boost performance on complex HSE tasks, we introduce the reasoning of experts, which simulates legal reasoning from multiple expert roles and perspectives.

\subsection{Dataset Instance Collection}
\benchmarkname~is designed to evaluate the capacities of LLMs in high-stakes, regulation-intensive contexts concerning health, safety, and the environment. 
To capture the diversity of real-world HSE contexts, we construct our benchmark using four representative data domains, specifically:

\begin{itemize}
    \item \textbf{Regulations}: HSE regulations from different countries and states. We manually collect data from the US, Canada, China, and Australia, and randomly sample portions of the regulations to prompt the LLM to generate regulation-based scenarios and questions.
    
    \item \textbf{Court Cases}: Real-world HSE-related court cases issued by courts in various regions. We primarily focus on cases from different states in Australia. Using the online website~\href{https://app.courtaid.ai/}{CourtAid}, we filter relevant cases based on HSE-related keywords. The full case documents are then sent to LLM to construct scenario-based questions and extract corresponding labels.
    
    \item \textbf{Safety Exams}: We collect various materials used in professional training, including those on HSE procedures and equipments, along with their corresponding labels. We then use an LLM to generate questions and extract the associated labels.
    
    \item \textbf{Fieldwork Videos}: We collect real-world instructional videos to evaluate hazard recognition and HSE compliance. These videos are then used to construct questions by an LLM. To enable evaluation with text-only LLMs, we convert the videos into textual descriptions.
\end{itemize}

In this way, the correctness of the labels for regulations, court cases, and safety exams is ensured, as the questions are directly generated from the corresponding sources. The labels for fieldwork videos and other sources are double-checked by experts to ensure accuracy and reliability.
These domains are selected to cover a broad spectrum of HSE scenarios, enabling a comprehensive evaluation of LLMs in knowledge understanding, prospective reasoning, and HSE compliance across diverse scenarios. The detailed sources of the data are listed in the appendix~\ref{sec:benchmark_details}.

\subsection{Dataset Taxonomy}

After collecting data from various sources, we adopt the IRAC framework, a structured reasoning schema from legal studies, to decompose the HSE compliance assessment task into four canonical steps of deductive reasoning, as follows:

\begin{itemize}
    \item \textbf{\textit{I}ssue Spotting}: This step aims to identify the core regulatory concern in the scenario and evaluate the ability of the LLM to uncover underlying causes beyond surface-level mentions.
    \item \textbf{Rule \textit{R}ecall}: This step is to identify the governing standard or regulation relevant to the issue spotted in the first step. It aims to evaluate the ability of the LLM to encode regulatory knowledge, understand domain terms, and identify relevant jurisdictions.
    \item \textbf{Rule \textit{A}pplication}: This step involves applying the recalled rule to the specific facts of the HSE scenario. It tests the ability of LLM to resolve ambiguity, handle rule conflicts, and reason under HSE contexts.
    \item \textbf{Rule \textit{C}onclusion}: This step involves drawing a final decision based on the applied rule and contextual facts. It tests the ability to produce outcomes that are both logically coherent and aligned with key HSE principles.
\end{itemize}

For each type of data source, we construct a legal reasoning chain along a unified cognitive IRAC framework. This framework operationalizes reasoning as a comprehensive taxonomy underlying our \benchmarkname. 
By integrating this step-by-step reasoning process with domain-specific scenarios, we are able to assess performance at different stages, uncover differences between model reasoning and legal reasoning, and assess whether models can internalize the legal reasoning process to enhance their performance in complex HSE contexts.

\noindent\textbf{Example}. We illustrate the IRAC framework using a concrete HSE scenario along with the corresponding question, options and labels. The examples are presented in Table~\ref{tab:irac_phase_example} of Appendix~\ref{sec:benchmark_details}.


\subsection{Benchmark Construction}

After curating high-quality HSE-relevant materials including regulations, court cases, safety exam, and fieldwork videos, we construct \benchmarkname~in two phases: (1) Initial prompt-based IRAC-style question generation to create diverse and structured legal scenarios, and (2) adversarial contextual enhancement to introduce distractors in the options, increase logical complexity and realism, and remove confidential information.

\subsubsection{Initial Prompt-Based Generation}
We construct prompt templates tailored to each domain for the initial prompt-based generation. Given a content excerpt, each prompt consists of: (1) a task description that guides the LLM to generalize the material-based scenario and generate IRAC-style questions; and (2) relevant materials, including the IRAC framework definition and the curated content. The detailed prompts for each type of data and the following generation are summarized in Appendix~\ref{sec:benchmark_details}.

These prompts are submitted to the LLM to generate a QA instance, which includes a scenario description, question, answer choices, correct label, explanation, and referenced rule. After generation, the outputs are manually reviewed and filtered during the early phases to calibrate LLM behavior.

\subsubsection{Adversarial Contextual Augmentation}
To further enhance the benchmark quality, we introduce the adversarial contextual augmentation phase.
We prompt the LLM with (1) a carefully constructed instruction template that guides it to rewrite each original HSE question into a more challenging and realistic format, and (2) the original question content, including scenario, options, and correct answer.
In particular, we design the instruction to emphasize logical complexity, plausible distractors, and confidential data. The prompt encourages the model to add complicating factors such as unclear responsibilities, conflicting information, or partial compliance, while preserving key facts and the original answer. This ensures the rewritten scenarios remain grounded in HSE scenarios.
Moreover, the resulting examples demand deeper critical thinking, forcing the model to go beyond surface cues and recognize underlying compliance gaps.
After the generation, two experts manually check each question.
A case study is conducted to validate the effectiveness of the augmentation process in ~Table~\ref{tab:case_study_4} in the Appendix~\ref{sec:mode_experimental_results}.

\begin{wraptable}{t}{0.5\columnwidth}
    \centering
    \vspace{-10mm}
    \caption{Data Statistics of \benchmarkname}
    \label{tab:cvqa-stats}
    \small
    \begin{tabular}{lr}
        \toprule
        No. of questions in regulation & 448 \\
        No. of questions in court case & 152 \\
        No. of questions in exam & 320 \\
        No. of questions in video & 100 \\
        Total number of IRAC questions & 1020 \\ \hline
        Avg. words per question in regulation & 60.24 \\
        Avg. words per question in court case & 111.76 \\
        Avg. words per question in exam & 92.49 \\
        Avg. words per question in video & 106.65 \\ \hline

        Avg. words per option in regulation & 10.99 \\
        Avg. words per option in court case & 6.56 \\
        Avg. words per option in exam & 9.73 \\
        Avg. words per option in video & 9.42 \\ \hline

        Avg. words per question in issue spotting & 89.07 \\
        Avg. words per question in rule recall & 91.94 \\
        Avg. words per question in rule application & 97.98 \\
        Avg. words per question in rule conclusion & 92.14 \\ \hline
        Avg. words per option in issue spotting & 7.69 \\
        Avg. words per option in rule recall & 7.21 \\
        Avg. words per option in rule application & 11.48 \\
        Avg. words per option in rule conclusion & 10.33 \\
        \bottomrule
    \end{tabular}
\vspace{-10mm}
\end{wraptable}

\subsection{Dataset Characteristics}

Table~\ref{tab:cvqa-stats} shows the statistics and the different sub-types of \benchmarkname. The \benchmarkname~dataset consists of 1020 IRAC-formatted questions with 448, 152, 320 and 100 questions sourced from regulations, court cases, exams, and videos, respectively. On average, questions from court cases (111.76 words) and videos (106.65 words) are longer than those from exams (92.49 words) and regulations (60.24 words). Across the four reasoning stages, rule application has the highest average question length (97.98 words) and option length (11.48 words), indicating more complex scenarios. In contrast, issue spotting and rule recall have shorter options (7.69 and 7.21 words, respectively), suggesting more focused decision points.

\subsection{Reasoning of Experts}
 
After a comprehensive evaluation of existing LLMs in \benchmarkname, we find that (1) although current LLMs demonstrate strong performance, their capabilities largely rely on surface-level semantic matching rather than structured reasoning based on legal knowledge, and (2) their native reasoning trace lacks the systematic legal reasoning required for rigorous HSE compliance assessment.

Motivated by these observations and inspired by real-world judicial decision-making, which often involves collaborative deliberation among jurors and judges, we propose a reasoning-of-expert (RoE) prompt, a simple but effective prompt for HSE compliance assessment by simulating the reasoning process of different experts. The prompt is designed as:

\textit{"This task involves evaluating an HSE legal or safety issue using multiple expert roles tailored to the case. Each expert provides an independent opinion using IRAC (Issue, Rule, Application, Conclusion) and step-by-step reasoning. A final judgment then synthesizes their insights into a justified, well-reasoned decision that balances regulatory compliance with practical safety."}

RoE is motivated by the idea of collaboration among LLMs which is widely used in the related literature~\cite{abdelnabi2024cooperation, grossmann2025power}. It prompts LLM to assume the role of multiple domain-specific experts. Each expert is instructed to generate an independent reasoning process based on the IRAC framework, grounded in their respective domain perspectives. Following the generation of multiple expert viewpoints, the model is prompted to synthesize these opinions. 
This process not only facilitates cross-validation among experts but also establishes a structured framework for reasoning. Experiments in Section~\ref{sec:evaluation} show that RoE achieves up to a 25.01\% improvement on DeepSeek-R1 as measured by accuracy compared to CoT.
A case study on the reasoning trace of RoE is illustrated in Table~\ref{tab:case_study_3} of Appendix~\ref{sec:mode_experimental_results}.
\section{Evaluation}
\label{sec:evaluation}
\subsection{Setup}
\noindent\textbf{Models}. We evaluate 12 state-of-the-art LLMs on the \benchmarkname, including both leading open-source and commercial models. These models span three categories: foundation models (Deepseek-V3, Qwen2.5-72B-Instruct, GPT-4o-mini, Llama-3.1-8B-Instruct, and Gemini-2.0-flash), reasoning models (Deepseek-R1, QwQ-32B, GPT-o3-mini, Claude 3.7-sonnet, and Gemini-2.5-pro-pre), and vision models (Llava-onevision-7B and Gemma-3-27B). We provide the necessary code to infer the model on \benchmarkname~Github Repository. 
Please note that questions in \benchmarkname~are knowledge-intensive, and search engines may be helpful. Most native large model APIs currently do not natively support internet search. Some models, such as Claude Sonnet and Gemini, can autonomously determine whether a search is needed. 
We do not integrate any external search engines across the board.
For video LLMs, we follow the existing method~\cite{bai2025qwen2,li2024llava} to extract frames from the video and use a leading multi-modal model, GPT-4o, to generate textual descriptions for the video.
The details of the evaluated models and the extraction prompts are summarized in the Appendix~\ref{sec:experimental_setup} and~\ref{sec:benchmark_details}. 

%

\noindent\textbf{Prompts}. To investigate the impact of different prompting strategies, following existing works, we evaluate 6 representative prompts. 
These include Zero-Shot, Few Shot, CoT, Zero-CoT (0-CoT), CoT with Self-Consistency (CoT+SC), and our Reasoning of Experts (RoE). The details of the construction of these prompts are summarized in the Appendix~\ref{sec:benchmark_details}.


\noindent\textbf{Metrics}. 
To comprehensively evaluate LLM performance, we adopt two metrics: Accuracy~\cite{fisher2009accuracy, wang2024efficient, wang2024missing} and AUC-ROC~\cite{narkhede2018understanding}. For the Accuracy metric, we provide the complete input, including the scenario, question, and all answer options, and prompt the LLM to select one option from the given choices. The selected answer is then directly compared with the ground-truth label to determine correctness.
For AUC-ROC, we input only the scenario and question, and obtain the free-form output of LLMs. We use BERT~\cite{devlin2019bert} to encode both the output and each candidate option, compute the cosine similarity between the output embedding and each option embedding, and treat similarity scores as confidence scores of LLM in that option. The AUC-ROC is then calculated to reflect the ability of LLM to assign higher confidence to the correct option compared to incorrect ones.

\subsection{Performance}
\subsubsection{Overall Results}

\begin{figure*}[t]
    \centering
    \begin{subfigure}[b]{0.47\textwidth}
        \includegraphics[width=\textwidth]{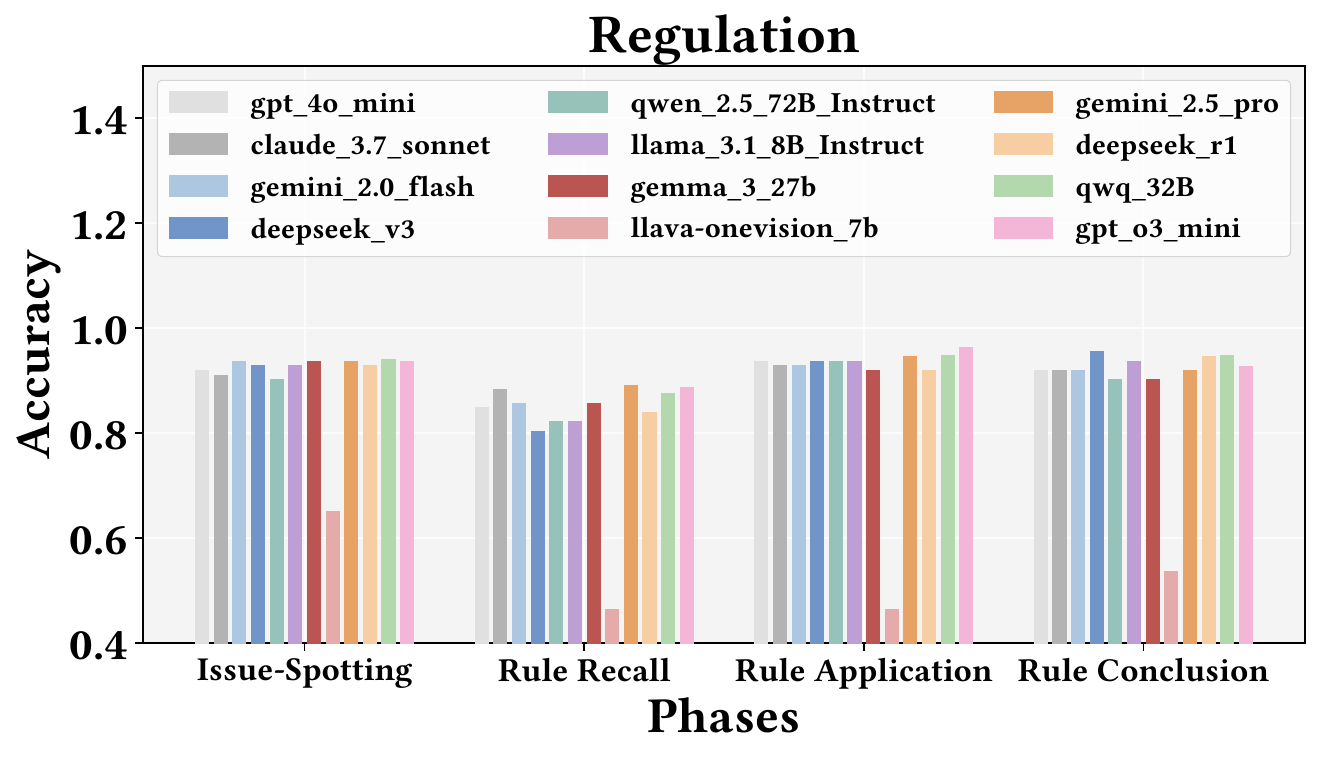}
        \vspace{-6mm}
        \caption{Results of regulation (Accuracy)}
    \end{subfigure}
    \hfill
    \begin{subfigure}[b]{0.47\textwidth}
        \includegraphics[width=\textwidth]{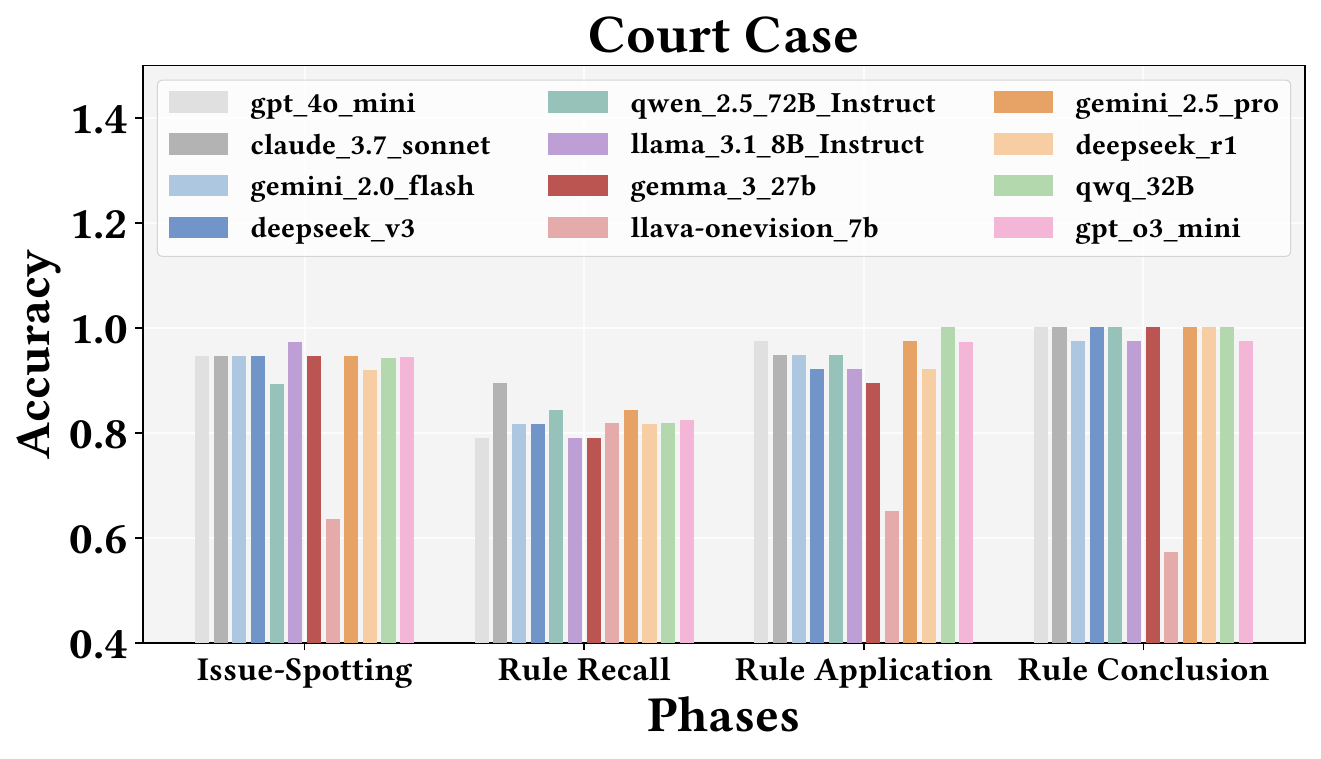}
        \vspace{-6mm}
        \caption{Results of court case (Accuracy)}
    \end{subfigure}
    \hfill
    \begin{subfigure}[b]{0.47\textwidth}
        \includegraphics[width=\textwidth]{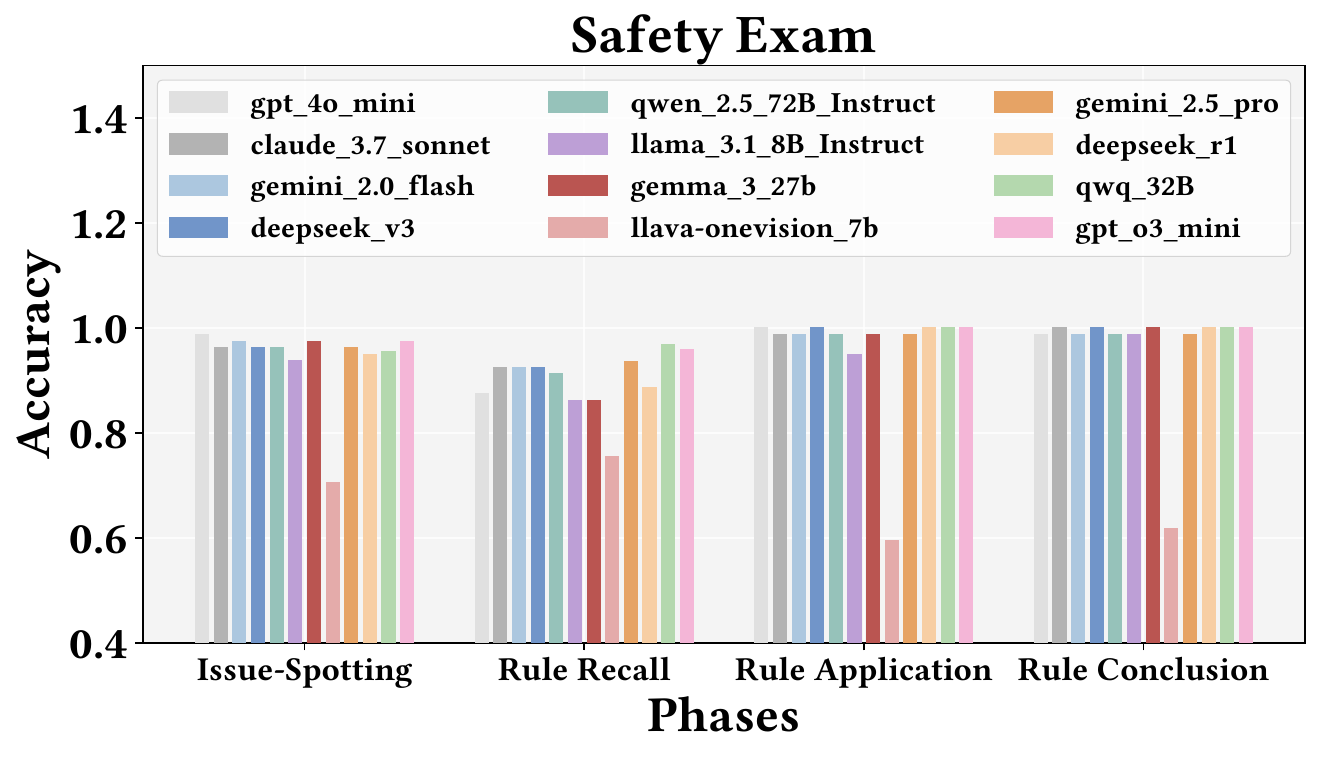}
        \vspace{-6mm}
        \caption{Results of safety exams (Accuracy)}
    \end{subfigure}
    \hfill
    \begin{subfigure}[b]{0.47\textwidth}
        \includegraphics[width=\textwidth]{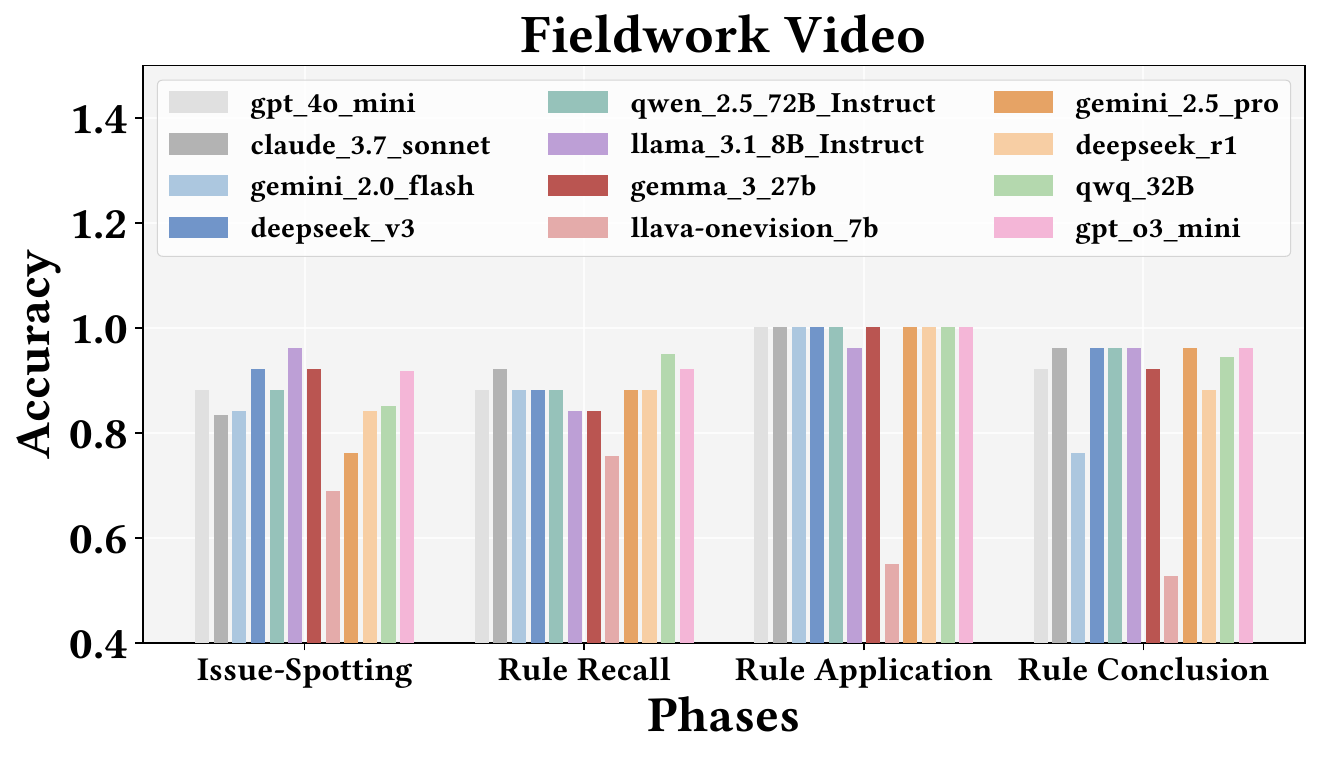}
        \vspace{-6mm}
        \caption{Results of fieldwork videos (Accuracy)}
    \end{subfigure}
    \hfill
    \begin{subfigure}[b]{0.47\textwidth}
        \includegraphics[width=\textwidth]{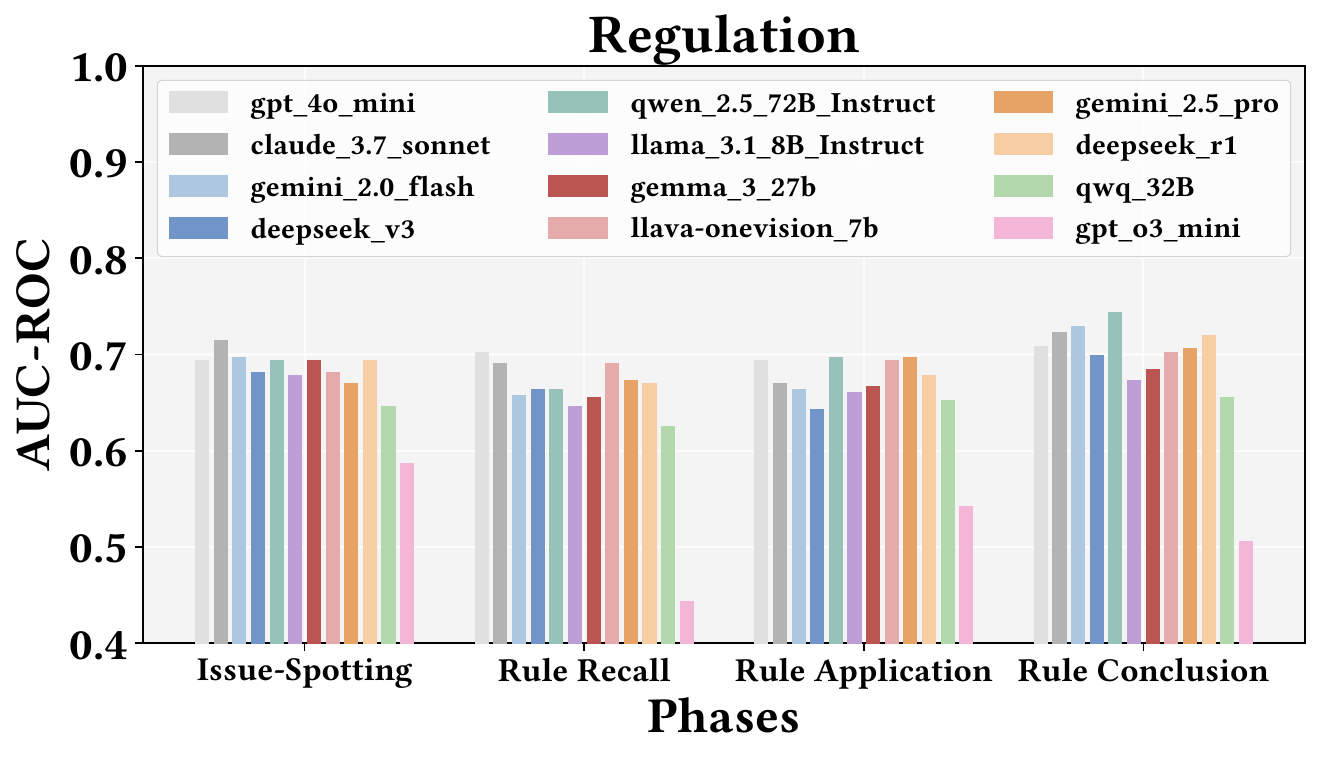}
        \vspace{-6mm}
        \caption{Results of regulation (AUC-ROC)}
    \end{subfigure}
    \hfill
    \begin{subfigure}[b]{0.47\textwidth}
        \includegraphics[width=\textwidth]{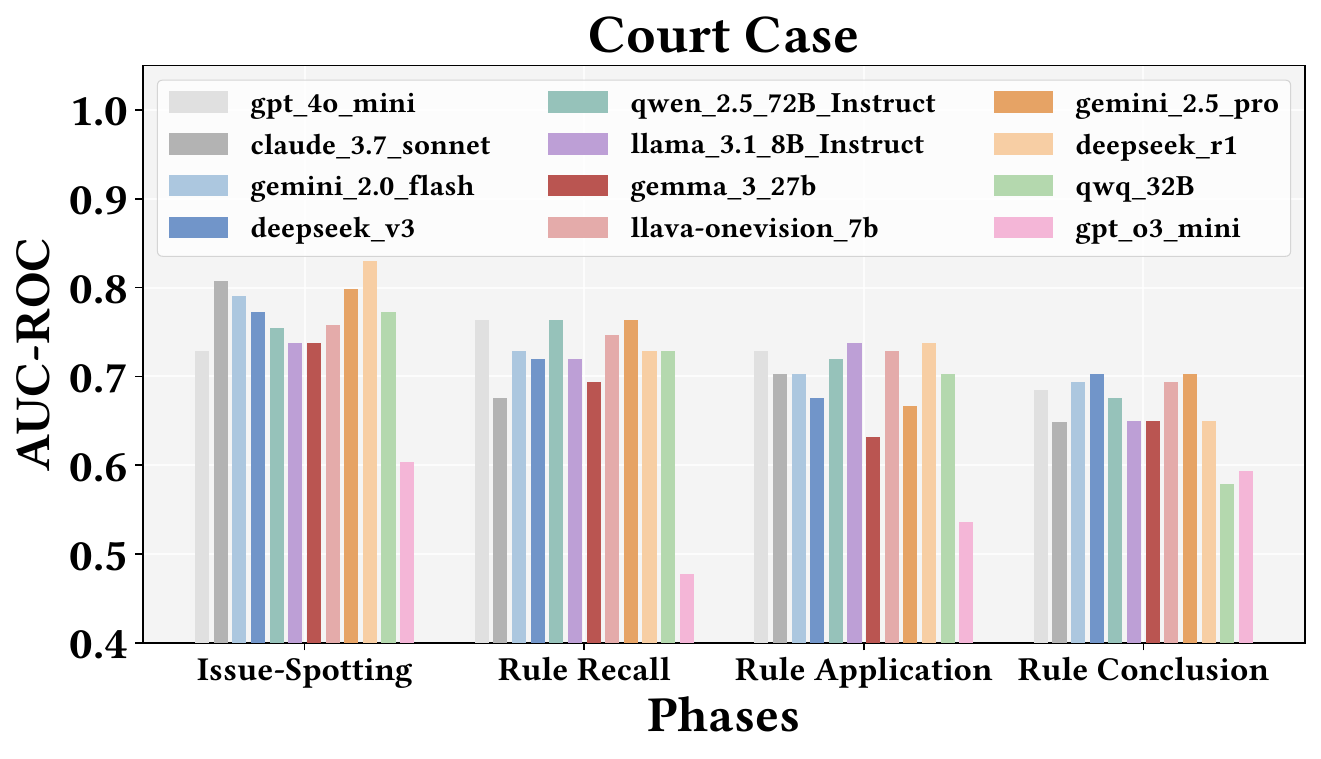}
        \vspace{-6mm}
        \caption{Results of court case (AUC-ROC)}
    \end{subfigure}
    \hfill
    \begin{subfigure}[b]{0.47\textwidth}
        \includegraphics[width=\textwidth]{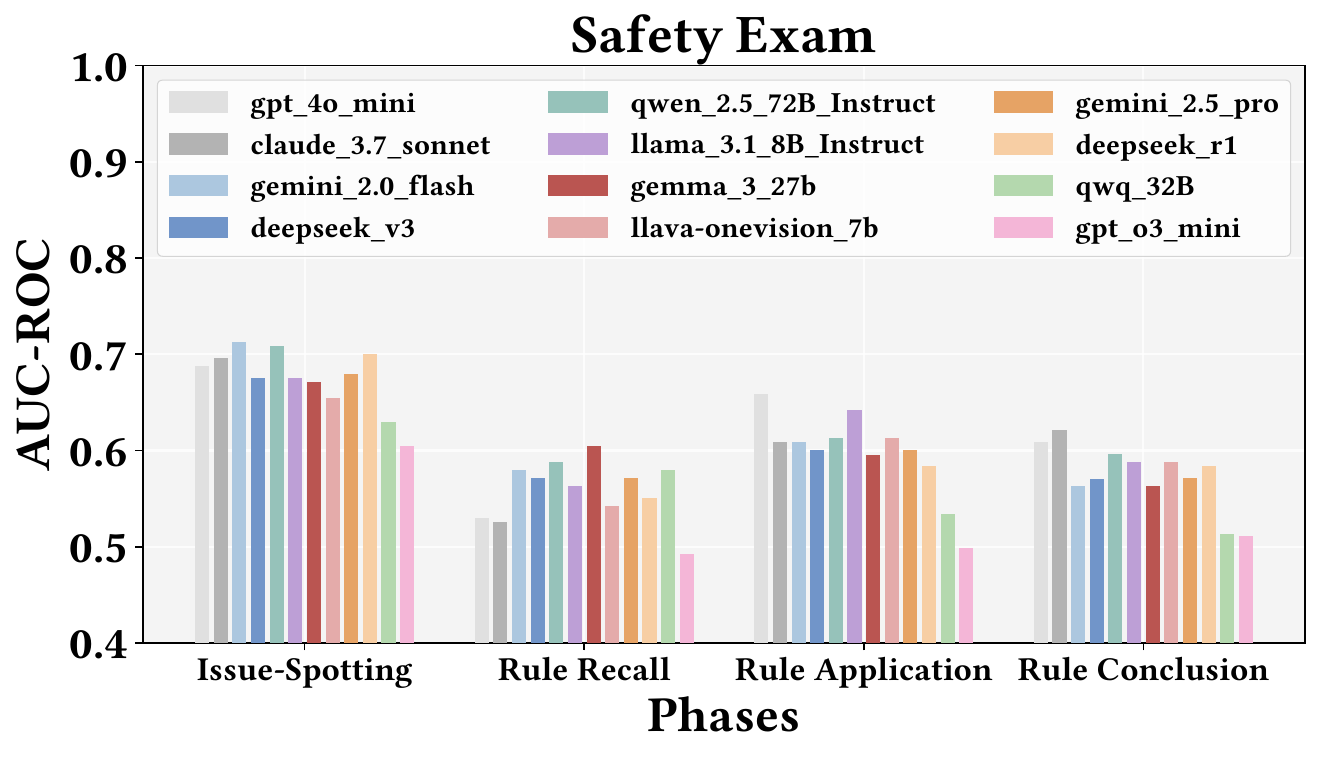}
        \vspace{-5mm}
        \caption{Results of safety exams (AUC-ROC)}
    \end{subfigure}
    \hfill
    \begin{subfigure}[b]{0.47\textwidth}
        \includegraphics[width=\textwidth]{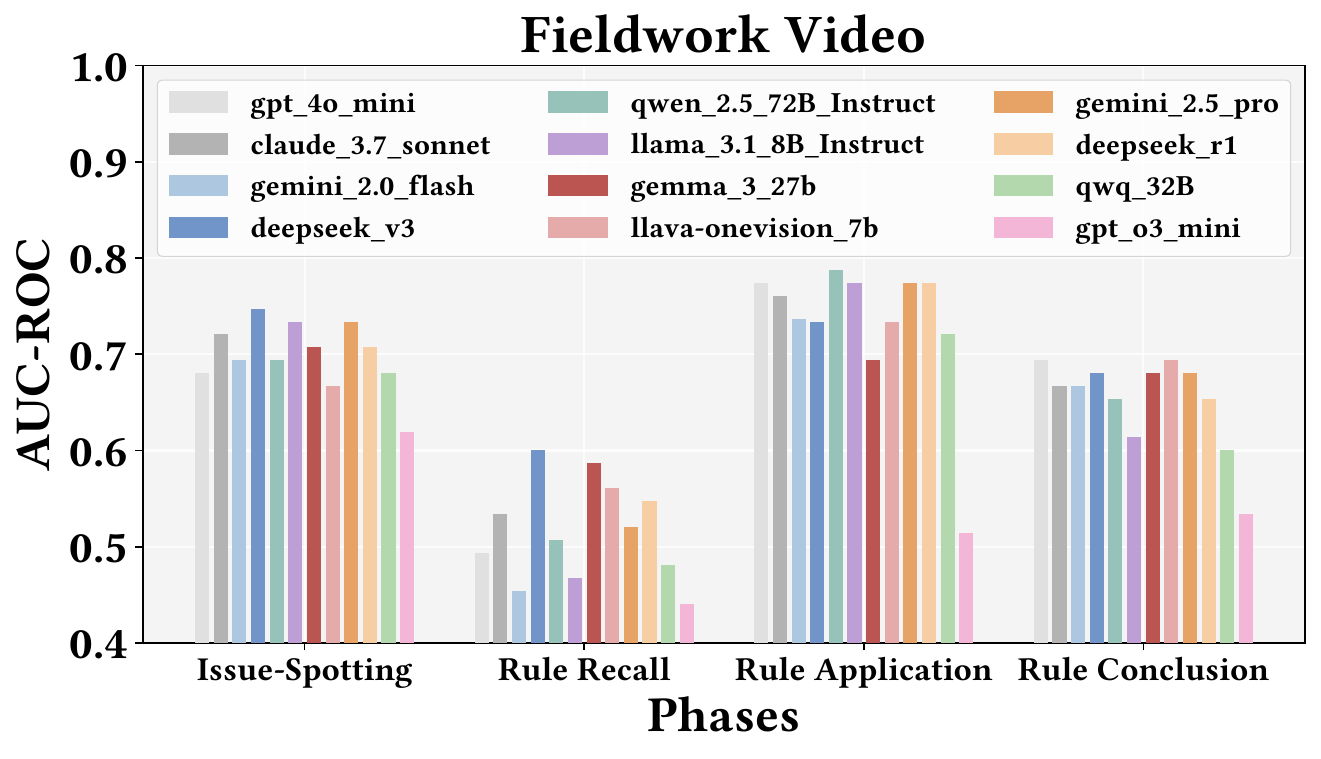}
        \vspace{-5mm}
        \caption{Results of fieldwork videos (AUC-ROC)}
    \end{subfigure}
    \vspace{-2mm}
    \caption{Overall results on different sources of data. Without options, LLMs show a sharp performance drop from $\sim90$\% accuracy to $\sim70$\% AUC-ROC, indicating a reliance on semantic matching between questions and options. Moreover, reasoning models fail to outperform foundation models. }
    \label{fig:overall_results}
    \vspace{-6mm}
\end{figure*}

Figure~\ref{fig:overall_results} presents the overall performance of various LLMs on \benchmarkname~, as measured by both accuracy and AUC-ROC. 

\noindent\textbf{Accuracy vs AUC-ROC}.
Comparing the results in terms of accuracy and AUC-ROC, we observe a significant performance drop of LLMs, with an average decrease of 24.73\% for all the models from different phases and data sources.
Moreover, GPT o3-mini shows the largest performance drop in rule application for the safety exam, with a sharp decline of 50.21\%. 
Note that under the same questions, settings, and single-choice format, AUC-ROC is typically higher than or equal to accuracy, as it captures the ability of LLM to rank the correct option highly even when the top choice is incorrect. Hence, the actual performance drop would be even greater if evaluated using the same metric. 
This disparity highlights the importance of providing explicit options to the LLM, and the LLM may achieve an outstanding single-choice performance by semantic matching in the HSE context~\cite{graydon2025examining}.
It also highlights how different input formats can elicit distinct reasoning patterns.
When options are provided alongside the scenario, the model tends to perform comparative reasoning between the given context and the candidate choices, leading to more confident and accurate decisions. In contrast, when only the scenario is presented, which mimics more realistic, open-ended query settings, the model relies on internal legal knowledge, which may result in less discriminative behavior and thus lower AUC-ROC scores. 

To further verify this finding, we conduct a case study on DeepSeek-R1, one of the state-of-the-art open-source reasoning models. The results are summarized in Table~\ref{tab:case_study_1} of the appendix~\ref{sec:mode_experimental_results}, clearly highlighting the contrastive thinking with options and heuristic thinking without options.
Note that the vision models exhibit different patterns: Gemma-3 performs better by accuracy, while Llava-onevision performs worse, possibly due to differences in textual encoders.



\noindent\textbf{Foundation Models vs Reasoning Models}. 
As highlighted in the figure, the reasoning models do not outperform non-reasoning models on \benchmarkname.
Moreover, the results show that the performance drop of reasoning models (29.72\%) measured by AUC-ROC compared to accuracy is significantly larger than that of foundation models (25.52\%).
These results may be attributed to several factors. 
Firstly, \benchmarkname~is designed to evaluate multiple dimensions of complex regulatory reasoning in HSE contexts, which requires not only comprehensive knowledge but also a nuanced understanding and application of domain-specific regulations. 
Secondly, the reasoning process in reasoning models is more likely to generate results from multiple viewpoints, which may not align with the exact target solution and thus harm performance as measured by AUC-ROC.
Thirdly, the reasoning patterns of current models are primarily designed for general logical processing, which may not align well with the legal reasoning and deep thinking required to handle complex HSE scenarios effectively.


To further verify these finding, we conduct another case study on DeepSeek-R1 and QwQ-32B. The results are summarized in Table~\ref{tab:case_study_2} of Appendix~\ref{sec:mode_experimental_results}, which clearly highlights the differing reasoning patterns across models. The results show that they tend to approach problems from diverse perspectives rather than following a structured, step-by-step method like the IRAC framework.

\noindent\textbf{Different Phases}.
We then explore the impact of different phases in the IRAC framework. Overall, performance of LLM tends to decline in the later phases. The average AUC-ROC of all evaluated LLMs for issue spotting, rule recall, rule application, and rule conclusion is 70.17\%, 60.75\%, 66.96\%, and 64.27\%, respectively. This trend suggests that as the task progresses from issue spotting to rule conclusion, the reasoning complexity increases.
Rule recall is particularly challenging, as it not only requires extensive domain knowledge but also a deeper understanding of the HSE context.
Moreover, the performance variance across models also widens in the later phases, further supporting this observation. The average AUC-ROC variance across all evaluated LLMs for issue spotting, rule recall, rule application, and rule conclusion is 14.73\%, 20.42\%, 19.73\%, and 15.77\%, respectively. Both the decline in average AUC-ROC and the increase in variance indicate that rule recall and latter phases pose greater challenges for LLMs.

\begin{figure*}[t]
    \centering
    \begin{subfigure}[b]{0.47\textwidth}
        \includegraphics[width=\textwidth]{ 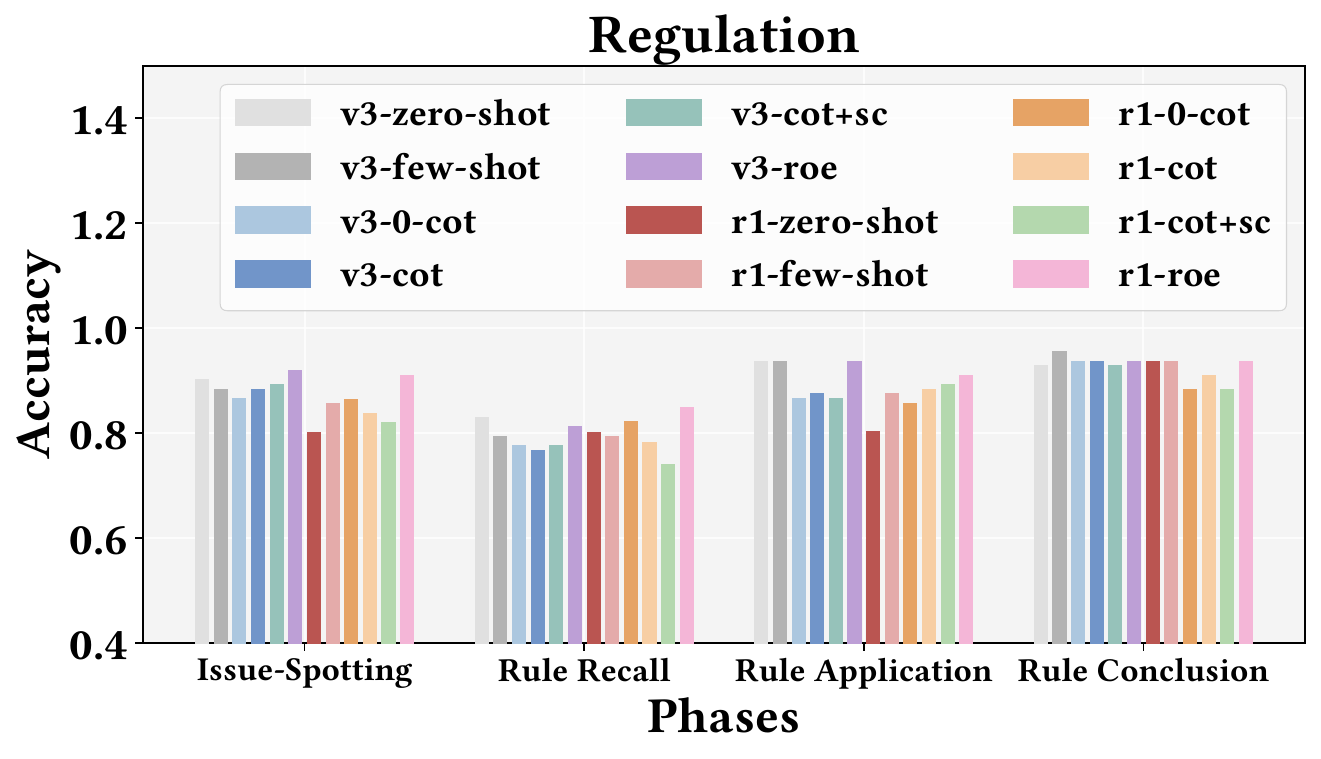}
        \vspace{-6mm}
        \caption{Results of regulation}
    \end{subfigure}
    \hfill
    \begin{subfigure}[b]{0.47\textwidth}
        \includegraphics[width=\textwidth]{ 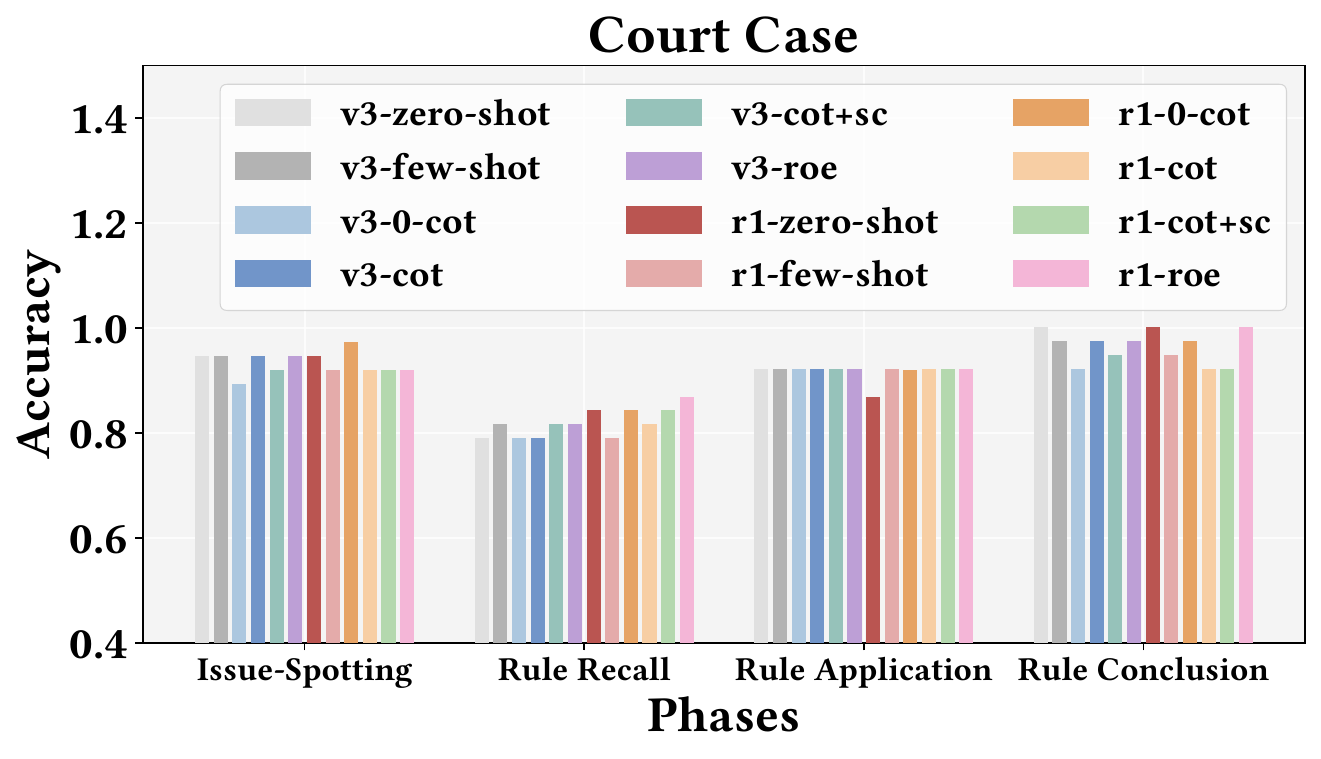}
        \vspace{-6mm}
        \caption{Results of court case}
    \end{subfigure}
    \hfill
    \begin{subfigure}[b]{0.47\textwidth}
        \includegraphics[width=\textwidth]{ 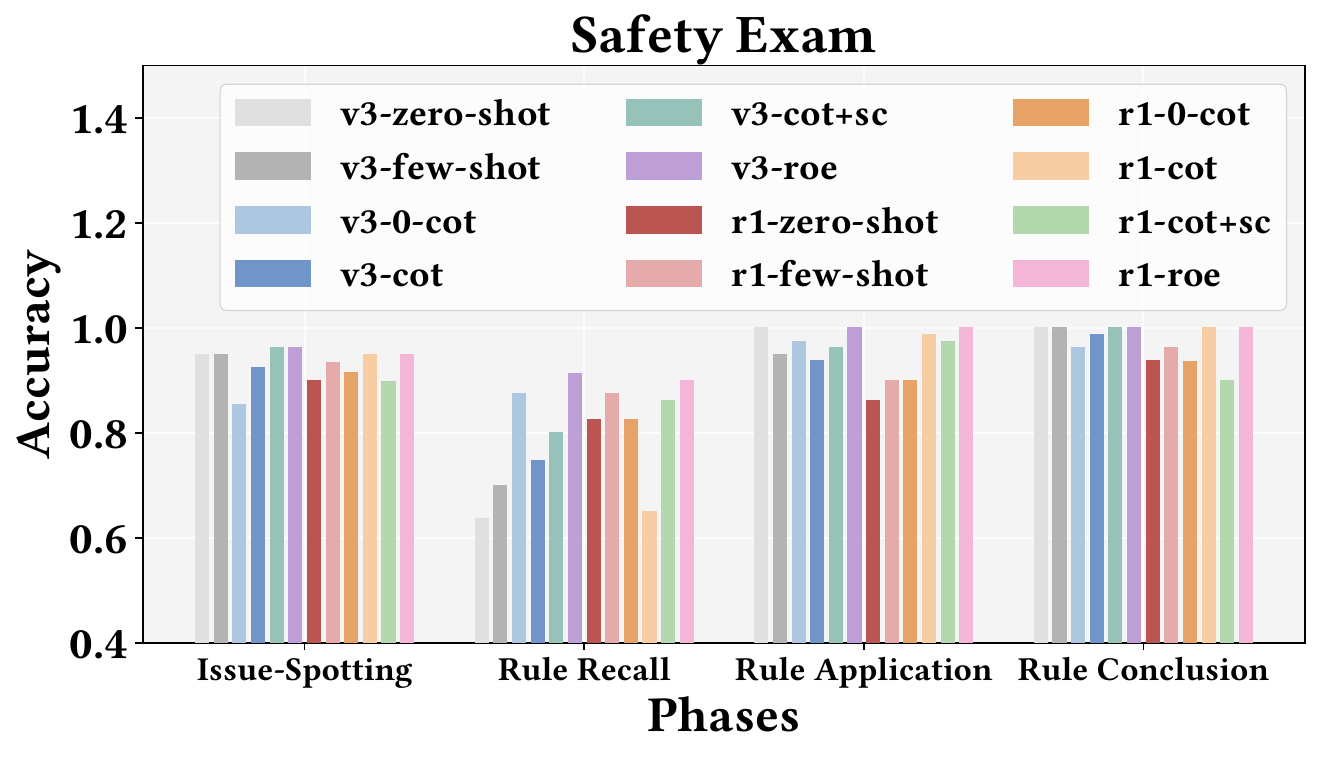}
        \vspace{-6mm}
        \caption{Results of safety exams}
    \end{subfigure}
    \hfill
    \begin{subfigure}[b]{0.47\textwidth}
        \includegraphics[width=\textwidth]{ 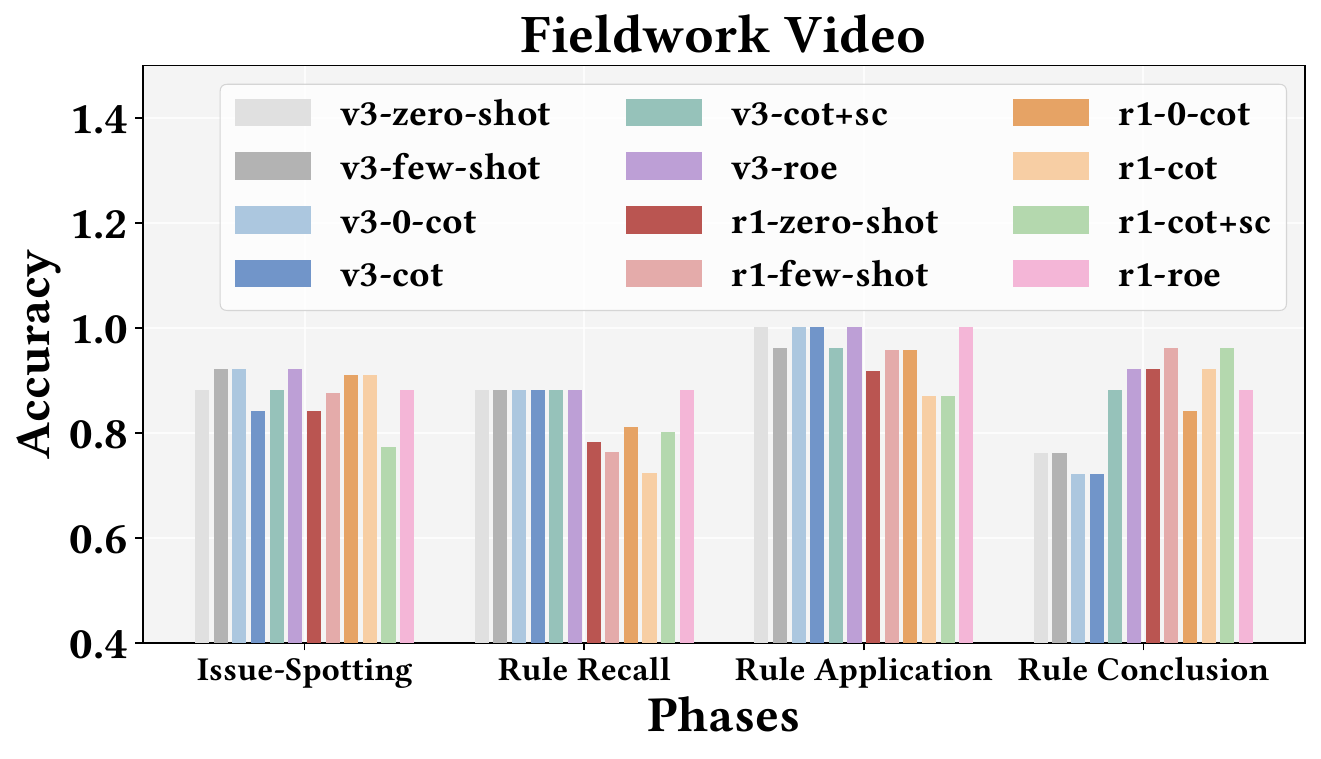}
        \vspace{-6mm}
        \caption{Results of fieldwork videos}
    \end{subfigure}
    \vspace{-2mm}
    \caption{Prompt strategies evaluation on both DeepSeek-V3 and DeepSeek-R1. Our RoE prompt significantly outperforms other prompting strategies on both LLMs.}
    \label{fig:prompt_evaluation}
    \vspace{-6mm}
\end{figure*}


\subsubsection{Prompt Evaluation Results}
In addition to evaluating various LLMs, we also assess different prompting strategies for both DeepSeek-V3 and DeepSeek-R1. The results are summarized in Figure~\ref{fig:prompt_evaluation}. As shown, the findings offer key insights into the effectiveness of these prompting strategies across various HSE contexts.
Firstly, CoT-based strategies (0-CoT, CoT, and CoT+SC) do not show significant improvements over naive strategies such as zero-shot and few-shot prompting. In some cases, they even perform worse, suggesting that simple CoT prompting may not be well-suited for HSE compliance assessment. Specifically, zero-shot, few-shot, 0-CoT, CoT, and CoT+SC achieve average accuracies of 89.76\%, 89.67\%, 88.47\%, 88.32\%, and 89.95\%, respectively, for DeepSeek-V3, and 87.40\%, 89.15\%, 88.90\%, 87.49\%, and 87.35\% for DeepSeek-R1.
Secondly, for reasoning-centric models, appropriate prompting strategies can lead to noticeable performance gains. This may be attributed to two factors: the relatively lower base performance of reasoning models and their stronger reasoning capacity, which enables them to adapt more effectively to different tasks through tailored prompts.
Notably, our proposed RoE strategy further boosts performance. It achieves average accuracy gains of 3.10\%, 3.19\%, 4.39\%, 4.54\%, and 2.91\% over zero-shot, few-shot, 0-CoT, CoT, and CoT+SC, respectively, for DeepSeek-V3, and 5.12\%, 3.37\%, 3.63\%, 5.03\%, and 5.18\% for DeepSeek-R1.
These results highlight the effectiveness of RoE, underscoring its strength in handling complex, multi-step reasoning tasks in HSE compliance assessment. 
Moreover, the consistent improvements across both DeepSeek-V3 and DeepSeek-R1 suggest the robustness of RoE. 
We also conduct a case study in Table~\ref{tab:case_study_3} of Appendix~\ref{sec:mode_experimental_results} to highlight the characteristics of the reasoning trace of RoE.
\vspace{-1mm}
\subsubsection{IRAC Pipeline Evaluation}
\begin{figure*}[t]
    \centering
    \begin{subfigure}[b]{0.47\textwidth}
        \includegraphics[width=\textwidth]{ 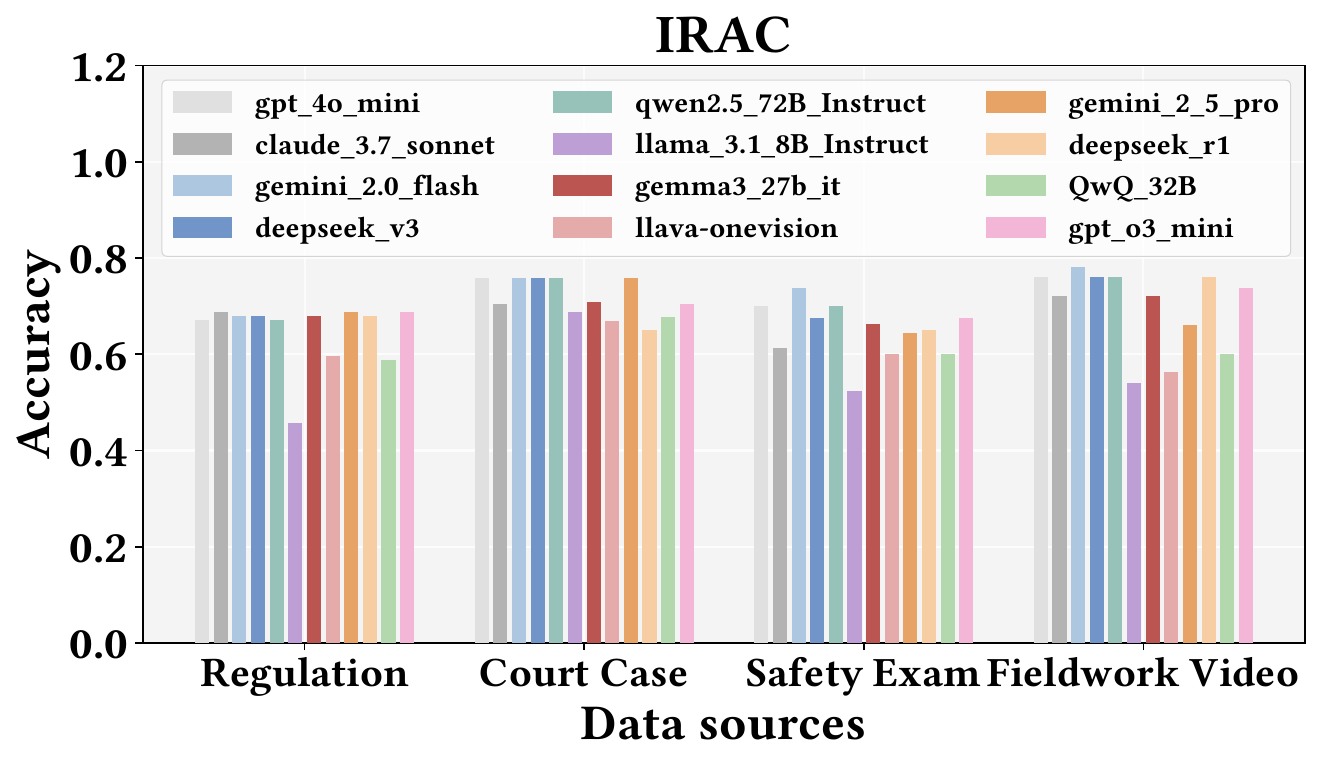}
        \vspace{-6mm}
        \caption{IRAC Results by accuracy}
    \end{subfigure}
    \hfill
    \begin{subfigure}[b]{0.47\textwidth}
        \includegraphics[width=\textwidth]{ 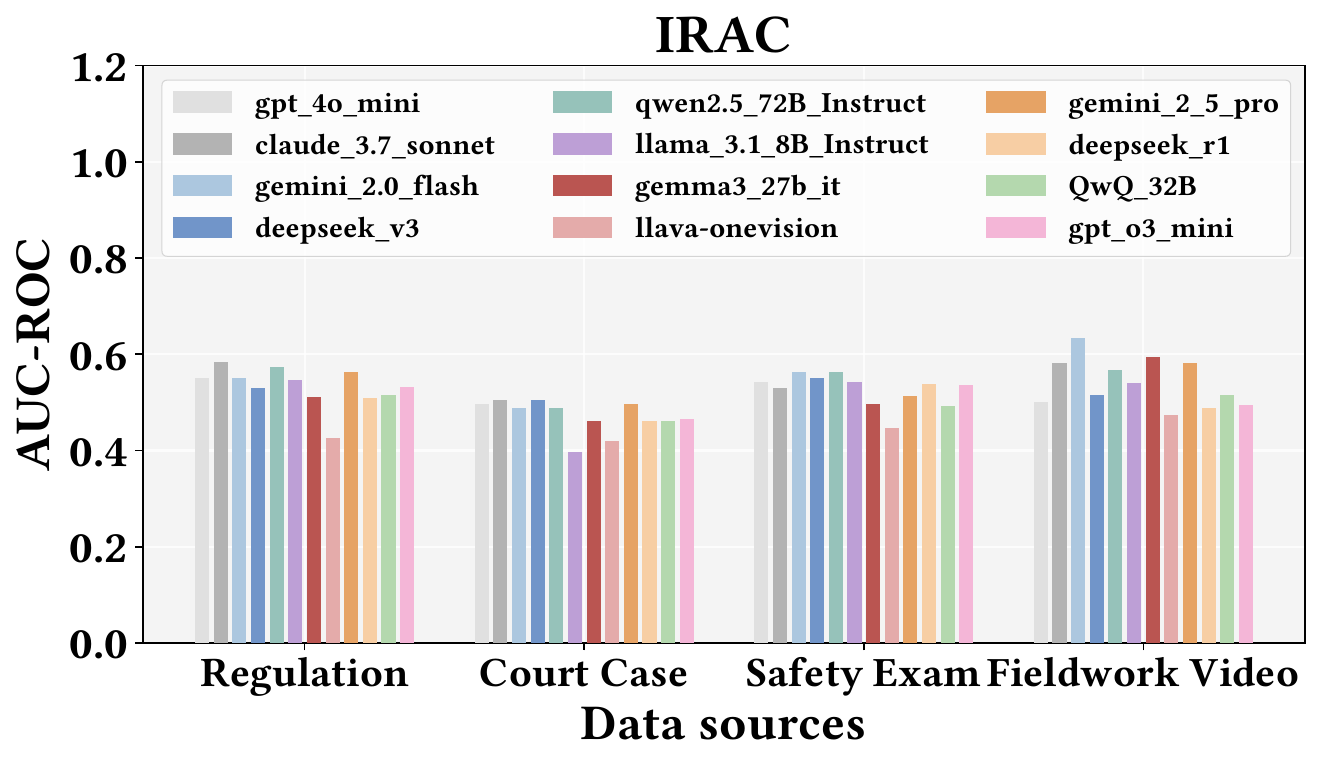}
        \vspace{-6mm}
        \caption{IRAC Results by AUC-ROC}
    \end{subfigure}
    \vspace{-2mm}
    \caption{IRAC reasoning evaluation. LLMs show a significant performance drop in the full IRAC pipeline, indicating difficulty in applying legal reasoning for HSE compliance. }
    \label{fig:irac_reasoning_evaluation}
    \vspace{-6mm}
\end{figure*}

Besides evaluating each phase independently, we also assess the entire IRAC pipeline in an integrated manner. Specifically, we collect the phases from the same scenario and construct composite questions that simultaneously test four phases in the IRAC. One of the answer options contains correct reasoning across all four phases, while the remaining three options include one or two reasoning errors at random positions.
The results are summarized in Figure~\ref{fig:irac_reasoning_evaluation}. As shown, performance drops significantly compared to the single-phase evaluations. For example, the average accuracy of Claude-3.7-Sonnet is 68.06\%, whereas its single-phase accuracy scores are 91.30\%, 90.58\%, 96.58\%, and 96.99\%, respectively.
Furthermore, the performance gap between different models becomes more pronounced, reflecting differences in their reasoning strategies. The average accuracy of DeepSeek-R1 is 68.42\%, while QwQ achieves 61.60\%.
As highlighted in the case study in Table~\ref{tab:case_study_2} in Appendix~\ref{sec:mode_experimental_results}, DeepSeek-R1 demonstrates a reasoning process that closely resembles legal logic, while QwQ tends to integrate perspectives from multiple roles. These observations further support our findings.




\vspace{-1mm}
\section{Conclusion}
\label{sec:conclusion}
In this work, we present HSE-Bench, the first comprehensive benchmark to evaluate LLM in HSE compliance assessment. Through over 1,000 carefully constructed questions and an IRAC-based evaluation framework, HSE-Bench reveals critical gaps in the legal reasoning capabilities of current LLMs. Our extensive empirical analysis shows that, while LLMs can perform reasonably well with semantic cues, they fall short in tasks requiring deep regulatory knowledge and structured reasoning. To address this, we introduce RoE, a novel prompting strategy that effectively enhances legal decision-making through expert role simulation and reasoning. We believe our findings shed light on the limitations of existing models and provide a solid foundation for future research in HSE.

\vspace{-2mm}

\paragraph{Limitations and Societal Impact.} HSE-Bench primarily focuses on the compliance assessment phase of HSE, leaving other critical stages in HSE, such as incident response, underexplored. While our work opens new directions for LLM applications, human oversight remains essential in high-stakes scenarios. Further research is needed to extend model evaluation and prompting strategies across the full HSE lifecycle.





\bibliographystyle{plain}
\bibliography{citation}

\newpage
\appendix

\section{Evaluated LLMs}
\label{sec:evaluated_llm}

In accordance with the NeurIPS Dataset and Benchmark Track guidelines, we publicly release the \benchmarkname~ dataset on Hugging Face: \url{https://huggingface.co/datasets/Joysouo/hse-bench}. 
Benchmark generation scripts, evaluation scripts, and metric computation codes are available in our GitHub repository: \url{https://github.com/mengqiwang1/hse-bench}. \benchmarkname~ is released under the Creative Commons Attribution License (CC BY 4.0).

We evaluate 12 LLMs, including foundation models, reasoning models, and vision-language models. Details such as model names, sizes, and API providers are listed in Table~\ref{tab:language_models}.


\begin{table}[h!]
\centering
\caption{List of evaluated LLMs in the experiments.}
\label{tab:language_models}
\begin{tabular}{l|l|c|c}
\hline
\textbf{Class} & \textbf{Model name} & \textbf{Model size} & \textbf{API provider} \\
\hline
\hline
\multirow{5}{*}{Foundation Models} 
  & Deepseek-V3 & 671B & Official platform\\
  & Qwen2.5-72B-Instruct & 72B & Nebius\\
  & GPT-4o-mini & --- & Official platform\\
  & Llama-3.1-8B-Instruct & 8B & Nebius\\
  & Gemini-2.0-flash & --- & Official platform\\
\hline
\multirow{5}{*}{Reasoning Models}
  & Deepseek-R1 & 671B & Official platform\\
  & QwQ-32B & 32B & Nebius\\
  & GPT-o3-mini & --- & Official platform\\
  & Claude 3.7-sonnet & --- & Official platform\\
  & Gemini-2.5-pro-pre & --- & Official platform\\
\hline
\multirow{2}{*}{Vision Models}
  & Llava-onevision-7b & 7B & Nebius\\
  & Gemma-3-27b & 27B & Nebius\\
\hline
\hline
\end{tabular}
\end{table}

\section{Experimental Setup}
\label{sec:experimental_setup}

\subsection{Evaluation prompts}
We evaluate five commonly used prompts from the literature and one newly proposed RoE prompt specifically designed for HSE compliance assessment. The details of the six prompts are listed below.

\noindent\textbf{1: Zero-shot prompting} evaluates the inherent capability of LLMs to generate responses without prior examples, serving as a baseline for assessing their raw encoded knowledge and reasoning skills. 

\noindent\textbf{2: Few-shot prompting} additionally provides the model with a limited number of examples (5 examples in our evaluation) to improve contextual understanding and response accuracy, particularly aiding in rule application tasks.
The template for zero-shot and few-shot prompts is as follows: 

\begin{lstlisting}
You are an expert in occupational health and safety regulations.
"k shot of different [Scenario,Question,Options,Answer]" # For few-shot prompt only
[Scenario and Question]: #scenario and question about the HSE context.
[Options]: # four options for the corresponding question.
Choose the best answer. Reply only with the letter A, B, C, or D. **Do NOT explain your choice in the content. Do NOT repeat the question or options.**
\end{lstlisting}

\noindent\textbf{3: CoT prompting} encourages step-by-step reasoning by explicitly providing the reasoning process as an example and guiding the model through intermediate reasoning stages before the final conclusion.

\noindent\textbf{4: 0-CoT prompting} explicitly instructs the LLM to reason systematically by the reasoning prompt.
The template for CoT and 0-CoT is as follows: 

\begin{lstlisting}
You are an expert in occupational health and safety regulations.
"[Scenario,Question,Options,ReasoningProcess,Answer]" # For CoT prompt only
[Scenario and Question]: #scenario and question about the HSE context.
[Options]: # four options for the corresponding question.
Choose the best answer. Reply only with the letter A, B, C, or D. **Do NOT explain your choice in the content. Do NOT repeat the question or options.**
"Let's think step by step:"
\end{lstlisting}

\noindent\textbf{5: CoT with Self-Consistency (CoT+SC)} enhances reliability by generating multiple independent reasoning paths and selecting the most consistent answer, thereby reducing errors due to stochastic variations in responses. We run the CoT 5 times and select the majority as the final answer.

\noindent\textbf{6: Reasoning of Experts (RoE)} simulates real-world legal decision-making processes, where multiple judges and jurors independently reason from different perspectives before reaching a final verdict. In this framework, the model assumes the roles of domain-specific experts. Each expert offers an independent judgment, which is then synthesized into a final decision, reflecting the compliance assessment procedures commonly used in HSE cases. This simple and effective template is as follows:

\begin{lstlisting}
You are an expert in occupational health and safety regulations.
[Scenario and Question]: #scenario and question about the HSE context.
[Options]: # four options for the corresponding question.
Choose the best answer. Reply only with the letter A, B, C, or D. **Do NOT explain your choice in the content. Do NOT repeat the question or options.**
"This task involves evaluating an HSE legal or safety issue using multiple expert roles tailored to the case. Each expert provides an independent opinion using IRAC (Issue, Rule, Application, Conclusion) and step-by-step reasoning. A final judgment then synthesizes their insights into a justified, well-reasoned decision that balances regulatory compliance with practical safety."
\end{lstlisting}

\subsection{Metric computation}
We evaluate model performance using two standard classification metrics: Accuracy and AUC-ROC (Area Under the Receiver Operating Characteristic Curve). Accuracy measures the proportion of correctly predicted labels over the total number of samples. It is computed as:
\begin{equation}
\text{Accuracy} = \frac{1}{N} \sum_{i=1}^{N} \mathbb{I}(\hat{y}_i = y_i)
\end{equation}
where $N$ is the total number of samples, $\hat{y}_i$ is the predicted label, $y_i$ is the ground-truth label, and $\mathbb{I}(\cdot)$ is the indicator function.

AUC-ROC evaluates the ability of LLM to rank the ground-truth option higher than negative options. The metric is computed as:
\begin{equation}
\text{AUC-ROC} = \frac{1}{|P||N|} \sum_{(i,j) \in P \times N} \mathbb{I}(s_i > s_j)
\end{equation}
where $P$ and $N$ denote the positive option and the sets of negative options respectively. $s_i$ is the predicted score (i.e., the cosine similarity between the option and response of LLM) for sample $i$. 

\section{Benchmark deatils}
\label{sec:benchmark_details}

\subsection{Data sources}
\label{sec:benchmark_details_data_sources}

To comprehensively construct the benchmark for HSE compliance assessment, we collect data from four data sources, including: regulation, court case, safety exam and fieldwork video.

\noindent\textbf{Regulation}. We manually collect regulation files from official work health and safety websites, including the following countries: (1) China \noindent(\url{https://www.chinajusticeobserver.com/law/x/work-safety-law-of-the-people-s-republic-of-china20210610/chn});
(2): Canada (\url{https://laws.justice.gc.ca/eng/acts/l-2/FullText.html}); 
(3) United States (\url{https://www.govinfo.gov/app/collection/cfr/2024/title29/subtitleA});
and the (4) Australia (\url{https://www.safeworkaustralia.gov.au/law-and-regulation/whs-regulators-and-workers-compensation-authorities-contact-information});  

We use the keywords of work health and safety and HSE to filter the files.
We then randomly sample segments of the regulation files and use them to prompt the LLM to generate regulation-based scenarios and corresponding questions. The prompt used for generating questions from regulation data is as follows:

\begin{lstlisting}
The details about this IRAC framework can be found here: "irac_rules". I now need to apply this framework to the HSE field in a way that challenges examinees to perform multi-layered legal analysis.

I have a regulation about HSE in location: "region_name", titled "regulation_name". The context of the regulation is: "original_text"

Based on these regulation contexts, generate scenario-based questions in {region_name} that adhere strictly to the IRAC framework. For each scenario, generate questions addressing four key components:

1. Issue Spotting: Identify the critical safety or legal issue presented in the scenario.
2. Rule Recall: State the relevant legal rule(s) or regulation(s) applicable to the identified issue.
3. Rule Application: Explain in detail how the legal rule applies to the given facts. The explanation must include:
   - Specific inferences drawn from the facts.
   - A detailed reasoning process that connects the facts to the rule.
   - A clear statement of the expected outcome based on these inferences.
4. Rule Conclusion: Present the final legal conclusion derived from the rule application.
For each of these four IRAC components, generate a multiple choice question with only one correct answer (labeled as m1_question)
For every question, provide: A correct answer (m1_answer).

Make sure to include the location in the scenario. 

Additionally, ensure adversarial generation of answer options by including distractor options: The distractor options should be plausible, referencing real-world regulatory elements but include subtle errors (e.g., incorrect clause numbers, misinterpretations of the rule, or irrelevant legal principles) to challenge examinees.
In summary, the generated questions should compel the examinee to:
   1. Identify nuanced safety issues.
   2. Recall and articulate specific legal rules.
   3. Apply these rules with detailed legal analysis and inference.
   4. Reach a reasoned legal conclusion supported by specific regulatory references.
   5. Critically evaluate adversarial answer options.
Generate the complete set of questions, answers for each IRAC component and each question format.
\end{lstlisting}

\noindent\textbf{Court Case}. We manually collect data from an online legal question-answering platform that contains a substantial number of court cases (\url{https://courtaid.ai/}). We then filter the cases using domain-relevant keywords including "HSE" and "WHS". Similar to the questions generated from the regulation, we also utilize LLM to generate question by the following template: 

\begin{lstlisting}
I have a court case about HSE in "region_name". I now need to apply the IRAC framework to generate test questions based on the court case. First, generalise the court case to a similar scenario that clearly states who do what at where. Second, we need four types of questions based on the IRAC framework, including issue spotting, rule recall, rule application, and rule conclusion. We require the option of issue spotting, which is the issue to be claimed in this scenario. The rule recall is the rule that is relevant to the issue. The rule application is the application of the rule to the scenario. The rule conclusion is the conclusion of the scenario based on the rule application. Third, we need to generate a single choice (m1_question) for each question type.
The details about this IRAC framework can be found here: "irac_rules"
The context of the court case is: "original_text".
\end{lstlisting}

\noindent\textbf{Safety Exam}. We manually collect data from the online HSE study platform \url{https://www.hsestudyguide.com/}. We randomly select 80 questions with their labels from the platform. We then use the following prompts for benchmark generation on safety exam.

\begin{lstlisting}
The details about this IRAC framework can be found here: "irac_rules". I now need to apply this framework to the HSE field in a way that challenges examinees to perform multi-layered legal analysis and critical evaluation.

I have a set of existing safety exam questions about HSE. The content of these questions is: {exam_content}. Note that the specific region applicable to these questions is indicated in the first line of the content.
Based on these safety exam questions, generate scenario-based questions that adhere strictly to the IRAC framework. For each scenario, generate questions addressing four key components:
 # for safety exam

I have a video about HSE. The description of this video is: "textual descriptions". Note that the specific region applicable to this scenario should be determined from the content.
Based on this video description, generate scenario-based questions that adhere strictly to the IRAC framework. For the scenario, generate questions addressing four key components: # for fieldwork video

1. Issue Spotting: Identify the critical safety or legal issue presented in the scenario.
2. Rule Recall: State the relevant legal rule(s) or regulation(s) applicable to the identified issue.
3. Rule Application: Explain in detail how the legal rule applies to the given facts. The explanation must include:
   - Specific inferences drawn from the facts.
   - A detailed reasoning process that connects the facts to the rule.
   - A clear statement of the expected outcome based on these inferences.
4. Rule Conclusion: Present the final legal conclusion derived from the rule application.

For each of these four IRAC components, generate a multiple-choice question with only one correct answer (labeled as m1_question).

For every question, provide:
- A correct answer (m1_answer).
- A comprehensive explanation (m1_explanation) that meets the following criteria:
   1. Correctness: The explanation must be factually accurate with no misstatements.
   2. Analysis: It must include detailed inferences and reasoning.
   3. Specificity: It must reference a specific, up-to-date HSE regulation (with name and clause number if applicable).
   4. Uniqueness: It must be unique and tailored to the scenario.

Additionally, include distractor options (e.g. trap option m1_trap_answer) with subtle errors. In the explanations, detail why the correct answer is valid and identify the misleading elements in distractors.

Generate the complete set of questions, answers, and explanations for each IRAC component based on the video description.
\end{lstlisting}

\noindent\textbf{Fieldwork Video}. We crawl data from YouTube (\url{https://www.youtube.com/}) using the keywords "construction site" and "HSE" to filter relevant videos. We select 9 high-quality videos and segment the last two into 30-second clips, resulting in a total of 25 videos: (1) Construction Workers' Master Work Skills; (2) Construction Tricks Idea; (3) Column Concreting Process Using a Lifting Mechanism; (4) Civil Construction Site Short Video; (5) ASMR Bricklaying — Bricklayer Satisfying Brickwork by Patterson Construction; (6) How Many Observations Can You Spot? (7) How to Build Concrete Footers for a House-2024 Edition (5:30-6:00); (8) How Iron Ore Mining Works; (9) Wonderful Bulldozer Construction: Hard Pushing Gravel, Building Road, Special Activities with Dump Trucks. 
Following existing work on multi-modal LLMs~\cite{li2024llava, bai2025qwen2}, we first extract frames from the video at 5-second intervals. The extracted frames are then sent to GPT-4o to generate textual descriptions by the following prompt. The prompt used for benchmark generation based on the video descriptions is listed above.
\begin{lstlisting}
The following video frames are extracted from the video at fixed time intervals (not consecutive frames). Please describe the scene, character actions, environmental changes, and other important details in the video.
\end{lstlisting}


\subsection{Implementation details of benchmark constructions}
\label{sec:benchmark_details_benchmark_construction}

To further enhance the quality of the generated questions, we design a two-step benchmark generation strategy. After the initial prompt-based generation, which directly constructs the source data and instruction to prompt the LLM for question generation, we introduce an adversarial contextual enhancement step. In this step, the initially generated questions are fed back into the LLM (GPT-4o) along with more detailed requirements. The newly designed prompts are as follows and the generated questions are then double-checked by domain experts.

\begin{lstlisting}
I am designing a benchmark question for HSE (Health, Safety, and Environment). You are help to rewrite a scenario-based question focused on issue spotting in a regulatory context.

Your task is to rewrite the **scenario and the final question sentence only**. The question tests whether the reader can identify the most relevant legal issue based on a realistic workplace situation.

Here is what you must do:
1. Write the scenario in clear, well-organized prose. It should sound like a real internal incident summary or audit finding. Use a **natural, professional, and concise tone**.
2. Do **not** mention any legal statute or regulation by name.
3. Make the issue **non-obvious**: include some context that may relate to incorrect options, creating plausible confusion.
4. Ensure the scenario includes at least one complicating detail (e.g., conflicting records, partial compliance, or unclear accountability).
5. Do **not** change the meaning of the correct answer, and do **not** modify the answer options.
6. Do not remove or alter any factual information such as the location, job roles, entities involved, or the outcome of the incident, but remove confidential information.

The original question is: "{question}"

The answer choices are: "{options}"

The correct answer is option "{answer}".

Please return only the rewritten scenario and final question sentence.
\end{lstlisting}

\subsection{Example of the benchmark taxonomy constructed using IRAC}
\label{sec:benchmark_details_example}

After applying our two-step benchmark generation strategy, we construct \benchmarkname, which follows the IRAC taxonomy for HSE compliance assessment.

The IRAC framework is a widely used method for legal analysis and structured reasoning~\cite{guha2023legalbench,stockmeyer2021legal}. It provides a systematic approach to breaking down complex HSE compliance assessment problems and constructing clear, logical arguments. IRAC stands for Issue, Rule, Application, and Conclusion, representing the four fundamental stages of analysis.

\begin{itemize}

\item \textbf{Issue Spotting}: This stage involves identifying the specific HSE compliance issue or problem raised by a set of facts. It requires the LLM to carefully read and critically analyze the underlying cause of the event based on the scenario-based description.

\item \textbf{Rule Recall}: In this phase, based on the spotted issue, relevant HSE compliance regulations are recalled from the encoded knowledge in LLM. A rule is a legal statement that specifies the necessary or sufficient conditions for a particular legal outcome to occur. The goal is to articulate the relevant regulations that govern the identified issue.

\item \textbf{Rule Application}: This stage involves applying the recalled legal rules to the facts of the case. It requires analytical reasoning to assess how the rules interact with the factual scenario and what implications arise.

\item \textbf{Rule Conclusion}: Finally, a conclusion is drawn based on the analysis. This conclusion should clearly answer the original legal question and reflect the reasoning developed in the previous stages.
    
\end{itemize}
We illustrate the generated benchmark and the IRAC framework in~Table~\ref{tab:irac_phase_example}. The correct options are highlighted in bold.
Each IRAC pipeline consists of four questions that share a similar scenario, with specific prompts and answer options.

\begin{table}
\centering
\caption{Examples of question in different IRAC phases (labels are highlighted in bold)}
\label{tab:irac_phase_example}

\begin{tabular}{m{1.5cm} m{12.0cm}}
\hline
\cellcolor{lightgray}{Types} & \cellcolor{lightgray}{Content} \\ \hline
\hline

Scenario & On 15 July 2022, BuildRight Ltd, a construction company operating in New South Wales, encountered a serious safety incident. During a project, Alex, a worker with limited experience, was tasked with operating a crane, a role typically requiring specialized training and certification. Despite Alex's lack of qualifications, the site foreman, under pressure to meet tight deadlines, instructed him to proceed. Unfortunately, Alex fell from a stack of construction beams, resulting in severe injuries. Complicating the situation, company records showed that Alex had been listed as a qualified operator, a discrepancy that was later attributed to administrative oversight. Records indicate that some safety protocols were followed, but there is ambiguity over who was responsible for verifying Alex's qualifications.  \\ \hline
Issue 

Spotting & What is the primary legal issue identified in this situation? 

\textbf{A}. Whether BuildRight Ltd provided adequate safety training to Alex.

\textbf{B}. Whether Alex had prior experience with cranes.

\textbf{C. Whether BuildRight Ltd ensured the health and safety of workers.}

\textbf{D}. Whether the construction beams were stacked correctly. \\ \hline

Rule 

Recall & What is the primary legal issue concerning workplace safety in this scenario?

\textbf{A. The Work Health and Safety Act 2011 (NSW).}

\textbf{B}. The Environmental Planning and Assessment Act.

\textbf{C}. Occupational Health and Safety Act 2000.

\textbf{D}. The Building Code of Australia.
\\ \hline
Rule 

Application & 

How should the legal rule be applied to BuildRight Ltd's situation?

\textbf{A. BuildRight Ltd did not assess the risks for inexperienced workers.}

\textbf{B}. BuildRight Ltd reviewed and updated its safety procedures regularly.

\textbf{C}. BuildRight Ltd provided all necessary safety gear but did not ensure it was used correctly.

\textbf{D}. BuildRight Ltd was unaware of the specific regulations that applied to their site.
\\ \hline
Rule 

Conclusion & 

What primary legal issue arises from BuildRight Ltd's handling of this situation?

\textbf{A. They violated safety regulations by not providing adequate training.}

\textbf{B}. They prioritized production over safety standards.

\textbf{C}. They lacked proper documentation for safety procedures.

\textbf{D}. They promoted a culture of safety among workers.
\\ \hline

\hline
\end{tabular}

\end{table}

\section{More experimental results}
\label{sec:mode_experimental_results}

\subsection{Additional results analysis}
\begin{figure*}[t]
    \centering
    \begin{subfigure}[b]{0.47\textwidth}
        \includegraphics[width=\textwidth]{ 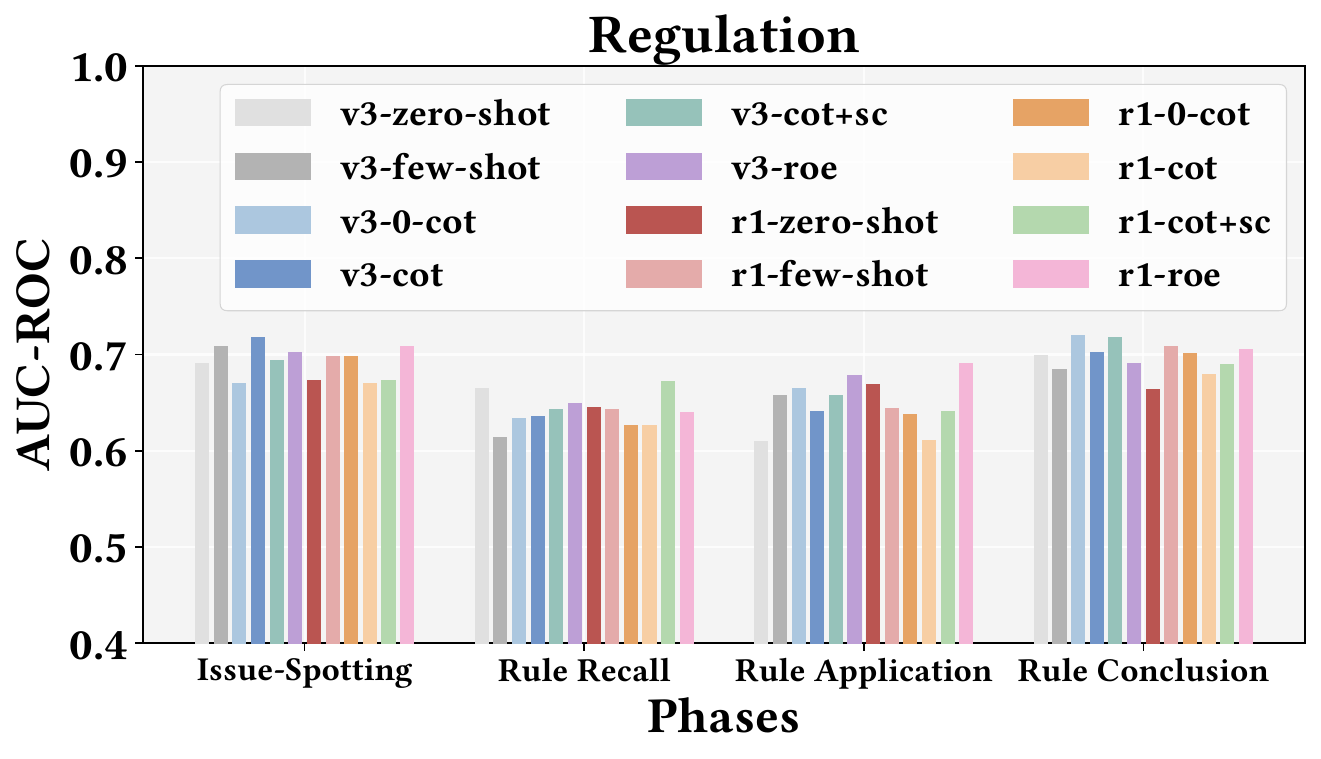}
        \caption{Results of regulation}
    \end{subfigure}
    \hfill
    \begin{subfigure}[b]{0.47\textwidth}
        \includegraphics[width=\textwidth]{ 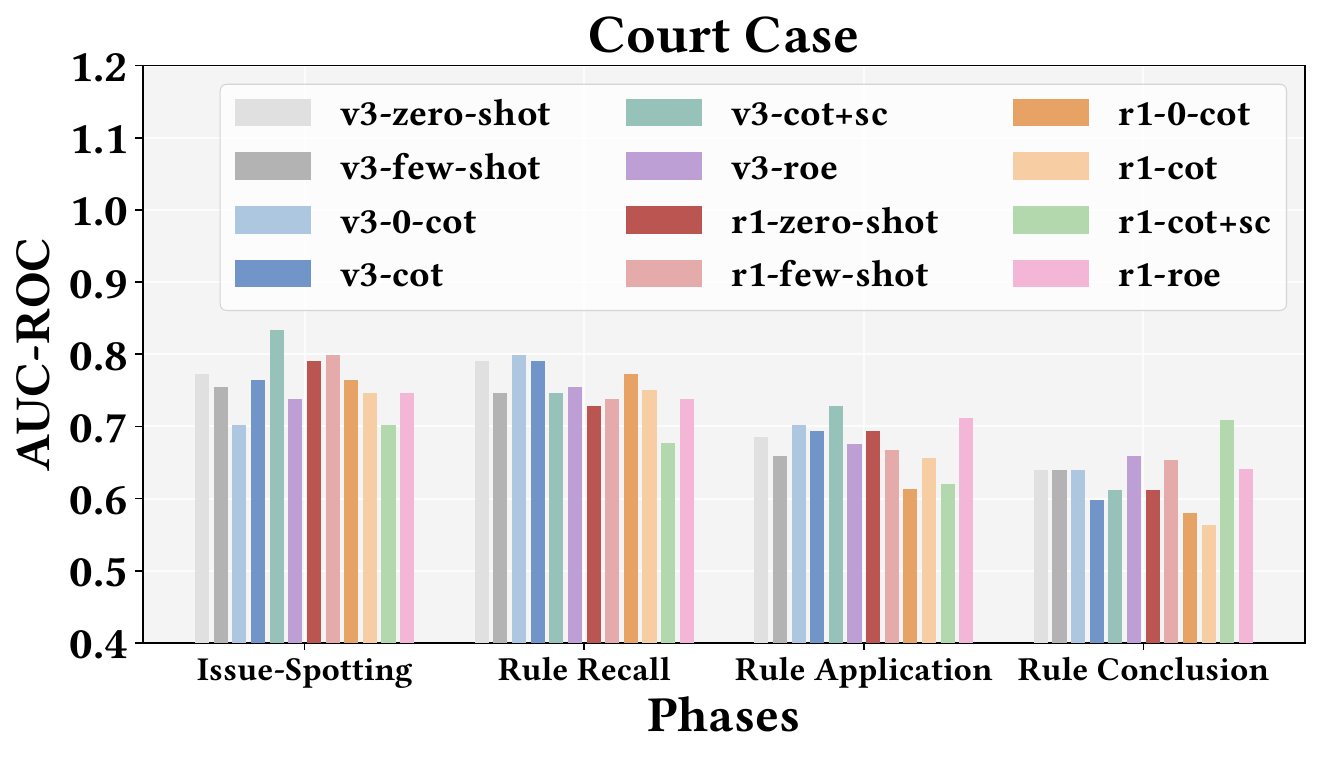}
        \caption{Results of court case}
    \end{subfigure}
    \hfill
    \begin{subfigure}[b]{0.47\textwidth}
        \includegraphics[width=\textwidth]{ 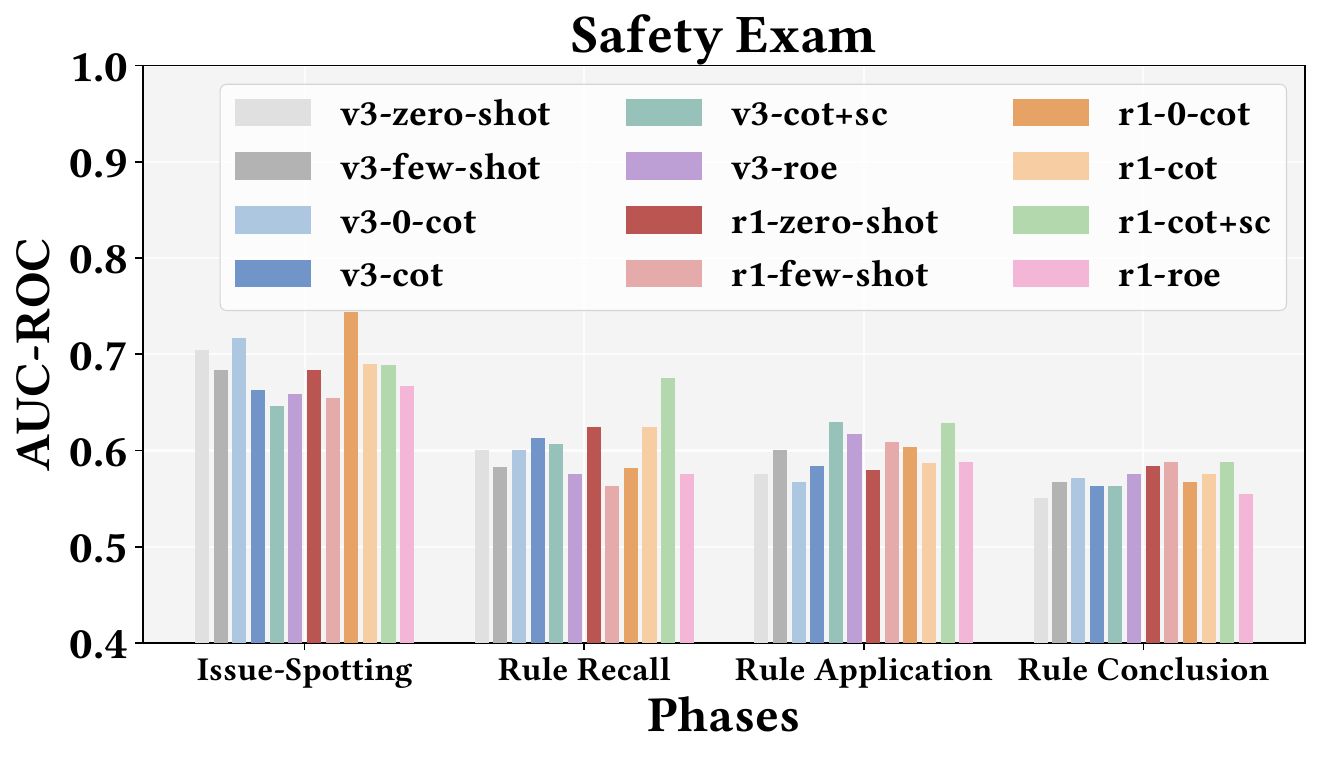}
        \caption{Results of safety exams}
    \end{subfigure}
    \hfill
    \begin{subfigure}[b]{0.47\textwidth}
        \includegraphics[width=\textwidth]{ 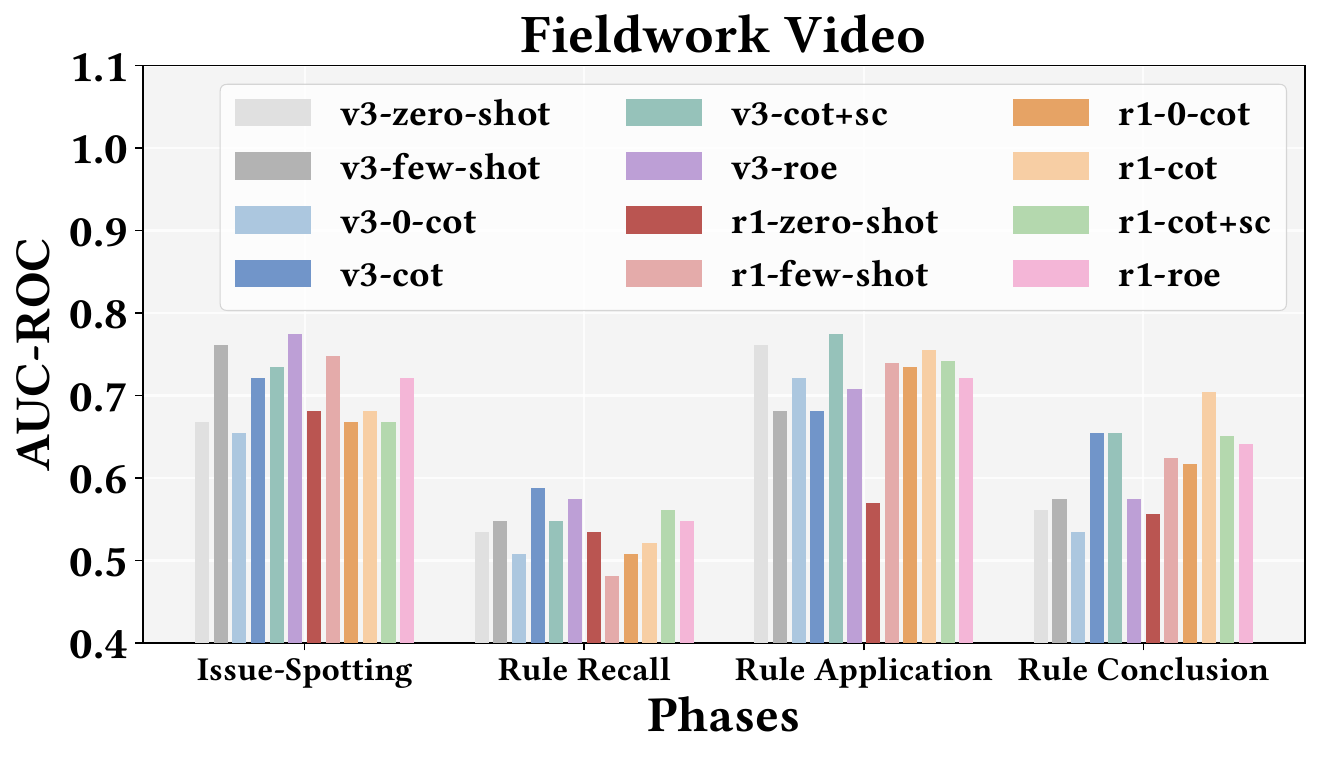}
        \caption{Results of fieldwork videos}
    \end{subfigure}
    \vspace{-2mm}
    \caption{Prompt strategies evaluation (AUC-ROC).}
    \label{fig:prompt_evaluation_AUC}
    \vspace{-6mm}
\end{figure*}

We also assess different prompting strategies as measured by AUC-ROC for both DeepSeek-V3 and DeepSeek-R1 to rigorously study their reasoning capabilities in HSE compliance assessment. The results are summarized in Figure~\ref{fig:prompt_evaluation_AUC}. 
As illustrated, the findings under AUC-ROC are similar to those of accuracy.
Firstly, CoT-based strategies (0-CoT, CoT, and CoT+SC) do not show significant improvements over naive strategies such as zero-shot and few-shot prompting. In some cases, they even perform worse, suggesting that simple CoT prompting may not be well-suited for HSE compliance assessment. Specifically, zero-shot, few-shot, 0-CoT, CoT, and CoT+SC achieve average AUC-ROC of 65.60\%, 65.32\%, 64.97\%, 66.24\%, and 67.36\%, respectively, for DeepSeek-V3, and 64.25\%, 65.92\%, 65.04\%, 65.19\%, and 66.10\% for DeepSeek-R1.
Secondly, for reasoning-centric models, appropriate prompting strategies can lead to more significant performance gains compared to those of foundation models.
Notably, our proposed RoE strategy further boosts the performance of CoT, with up to 6.07\% when tested on DeepSeek-V3 and up to 8.03\% when tested on DeepSeek-R1.
These results highlight the effectiveness of RoE prompting, underscoring its strength in handling complex, multi-step reasoning tasks in HSE compliance assessment.

\subsection{Case study}

We conduct four case studies to further validate our findings and assess the reasoning trace of LLMs. 

In the first case study shown in Table~\ref{tab:case_study_1}, we examine the reasoning traces and validate the finding that the capabilities of the LLM largely rely on semantic matching between the question and the options, rather than principled legal reasoning grounded in the underlying HSE compliance context. We conduct experiment on DeepSeek-R1.
As shown in the table, when provided with the question without options, the LLM primarily recalls relevant rules before analyzing the scenario. It identifies applicable legal principles, explains the incident, discusses the implications of inconsistent labeling and storage, and reasons through the legal status of the warehouse. Based on this reasoning pipeline, the LLM concludes the primary legal issue.
In contrast, the output for LLM with options adopts a more structured reasoning approach by first analyzing each option in the context of the scenario. It then compares options to distinguish and select the most appropriate one. It evaluates each option (A to D), eliminates less relevant ones, and highlights subtle distinctions. It then compares the remaining choices and selects the most appropriate answer based on legal principles and the scenario.

In the second case study shown in Table~\ref{tab:case_study_2}, we examine the reasoning traces of both DeepSeek-R1 and QwQ-32B and validate the observation that existing LLMs struggle to effectively apply the critical legal reasoning required for HSE compliance assessment. As shown in the table, both DeepSeek-R1 and QwQ-32B analyze the incident from different perspectives, such as that of the employer, and recall high-level legal rules. However, they lack a rigorous and specific legal reasoning pipeline, which leads to inaccurate conclusions.

In the third case study shown in Table~\ref{tab:case_study_3}, we illustrate the reasoning trace after using RoE as the prompt. As highlighted in the table, the LLM adopts a more structured reasoning process compared to the original version in Table~\ref{tab:case_study_2}, leading to a more accurate outcome.

In the fourth case study shown in Table~\ref{tab:case_study_4}, we illustrate the augmented question by our adversarial contextual augmentation process and demonstrate the effectiveness of this process.

\begin{table}
\centering
\caption{A case study highlights the differing reasoning traces of the LLM when options are provided versus omitted, validating that the capabilities of LLM rely largely on semantic matching rather than principled reasoning grounded in the underlying HSE compliance context.}
\label{tab:case_study_1}

\begin{tabular}{m{1.3cm} m{12.0cm}}
\hline
\cellcolor{lightgray}{Types} & \cellcolor{lightgray}{Content} \\ \hline
\hline
Questions & During an internal review at a facility in Sydney, NSW, operated by SafeChem Pty Ltd, it was discovered that an employee named John sustained injuries following a chemical leak. The incident was traced back to inconsistencies in labelling and storage practices. Complicating the issue, the warehouse where the leak occurred is not typically classified as a traditional workplace, which raised questions about the applicability of certain safety protocols. What is the primary legal issue identified in this situation? 

\textbf{A}. The general concept of workplace safety in unconventional spaces.

\textbf{B. The intricate handling and storage protocols of dangerous substances.}

\textbf{C}. The adequacy of employee training programs in handling emergencies.

\textbf{D}. The effectiveness of existing fire safety measures in non-traditional workplaces. \\ \hline

LLM Reasoning Trace (No Options) & Okay, so there's this scenario where an employee named John got injured because of a chemical leak at a SafeChem facility in Sydney. The problem was due to inconsistent labeling and storage. Also, the warehouse where it happened isn't usually classified as a traditional workplace. The question is about the primary legal issue here.

{\color{orange}First, I need to recall Australian workplace safety laws.} I know that Work Health and Safety (WHS) laws apply in NSW...

{\color{orange}The incident involved a chemical leak caused by inconsistent labeling and storage.} So, improper labeling might mean that workers didn’t know the risks, leading to incorrect handling. Storage issues could mean incompatible chemicals were stored together or not secured properly, causing a leak.

Now, the warehouse not being a traditional workplace complicates things. But the WHS Act defines a workplace broadly...

{\color{orange}So the primary legal issue is whether the warehouse is considered a workplace under WHS laws, thereby obligating SafeChem to maintain proper labeling, storage, and safety protocols.} If the warehouse is a workplace, then the company failed its duty of care by not ensuring correct practices, leading to John’s injury. The inconsistency in labeling and storage directly points to breaches of WHS regulations, like the Hazardous Chemicals Regulations, which require correct labeling and Safety Data Sheets. The confusion about the warehouse’s classification might be the company’s defense, but legally, if work is performed there, it’s a workplace. Therefore, the main issue is the application of WHS obligations to a non-traditional workplace.
 \\ \hline
LLM Reasoning Trace (With Options) & Okay, so I need to figure out the primary legal issue here. Let's start by breaking down the scenario. The incident happened at a facility in Sydney operated by SafeChem Pty Ltd. An employee, John, got hurt because of a chemical leak. The cause was inconsistencies in labeling and storage practices. Also, the warehouse where this happened isn't typically classified as a traditional workplace, which makes people question which safety protocols apply.

{\color{orange}First, the options are A through D. Let me go through each one.}

Option A: The general concept of workplace safety in unconventional spaces. The warehouse is not a traditional workplace, so maybe the issue is whether the usual workplace safety rules apply here...... 

Option B: Intricate handling and storage protocols of dangerous substances. The leak was due to labeling and storage issues. So maybe the problem is that they didn't follow proper procedures for handling dangerous chemicals...

Option C: Adequacy of employee training. The question mentions inconsistencies in labeling and storage, not directly about training. Unless poor training led to the incorrect labeling and storage... 

Option D: Fire safety measures in non-traditional workplaces. The incident is a chemical leak, not a fire. Unless the leak could lead to fire risks, but the question doesn't mention fire. So D might not be the primary issue here.

{\color{orange}So between A and B, ...., So maybe the answer is B. Because even if the space is non-traditional, once it's a workplace, the protocols for dangerous substances apply. The inconsistencies in labeling and storage are part of those protocols, so the primary legal issue is B.}....
\\ \hline

\hline
\end{tabular}

\end{table}
\begin{table}
\centering
\caption{A case study highlights the differing reasoning traces of DeepSeek-R1 and QwQ, supporting the observation that they struggle to effectively apply the critical legal reasoning required for effective HSE compliance assessment. }
\label{tab:case_study_2}

\begin{tabular}{m{1.3cm} m{12.2cm}}
\hline
\cellcolor{lightgray}{Types} & \cellcolor{lightgray}{Content} \\ \hline
\hline
Questions & At a drilling site in Queensland, an incident occurred involving Jacob Kilby, a driller for Saxon Energy Services, who operated the Iron Roughneck in a manner that resulted in a fatal accident involving his colleague, Gareth Dodunski. Despite existing safety protocols, Saxon Energy Services faces scrutiny over their accountability for Jacob's actions, as well as questions about Carl Thomas, another operator, who did not update the safety management plan after a similar incident at another location. Considering these factors, what conclusion can be drawn regarding Saxon Energy's liability? (Evaluation without options)

\textbf{A}. Saxon Energy is not liable at all.

\textbf{B. Saxon Energy is liable due to Kilby's actions being within his employment scope.}

\textbf{C}. Saxon Energy's liability is negated by contractual clauses.

\textbf{D}. Saxon Energy is liable only if Kilby was unsupervised. \\ \hline

DeepSeek-R1

Reasoning Trace & Okay, so I need to figure out Saxon Energy Services' liability in this fatal accident involving Jacob Kilby and Gareth Dodunski. Let me start by breaking down the information given.

{\color{orange}{First, Jacob Kilby}}, a driller for Saxon, operated the Iron Roughneck in a way that caused a fatal accident. There were existing safety protocols, but apparently, they weren't enough to prevent this....

I remember that in legal terms, employers can be held liable for the actions of their employees under the principle of vicarious liability...

{\color{orange}{Then there's Carl Thomas}} not updating the safety plan after a similar incident elsewhere. That seems important. If there was a prior incident and the safety plan wasn't updated, Saxon might be seen as negligent in maintaining a safe work environment....

Also, existing safety protocols were in place, but the accident still happened. The question is whether those protocols were adequate and properly enforced. If the protocols were insufficient...

In some jurisdictions, employers are required to ensure a safe workplace, which includes proper training, equipment maintenance, and updating safety plans...

Also, maybe there's a question about supervision...

{\color{orange}{So putting it all together: Saxon could be liable both vicariously for Jacob's actions if he was acting within his job duties, and directly for their own negligence in not updating the safety plan after the prior incident.}}
 \\ \hline
QwQ-32B Reasoning Trace & Okay, so I need to figure out what conclusion can be drawn regarding Saxon Energy's liability in this incident. Let me start by breaking down the information given.

{\color{orange}{First, there's Jacob Kilby}}, a driller for Saxon Energy Services, who was operating the Iron Roughneck, which is some kind of equipment. His operation....

{\color{orange}{Then there's scrutiny over Saxon Energy's accountability}} for Jacob's actions. That makes me think about employer liability. Employers are usually responsible for their employees' actions if those actions occur within the scope of their employment....

Hmm, so Carl didn't update the safety plan after a prior incident. That seems like a failure in safety management. If they had a prior incident and didn't update their protocols, that could indicate negligence on the part of the company...

I should consider the legal principles here. In employer liability, there's vicarious liability, where the employer is responsible for the employee's actions...

The fact that Carl didn't update the safety management plan after a similar incident suggests that the company didn't learn from past mistakes...

Also, safety protocols existed, but maybe they weren't adequate or were not properly enforced...

Another angle is whether Jacob's actions were a result of inadequate training or supervision...

Additionally, the fact that there was a prior similar incident at another location that wasn't addressed by updating the safety plan might mean that the company had a pattern of negligence....

{\color{orange}{So putting this all together, the conclusion would probably be that Saxon Energy Services is likely liable because they didn't update their safety protocols after a prior incident (Carl's failure), which contributed to the current accident. }}
\\ \hline

\hline
\end{tabular}

\end{table}
\begin{table}
\centering
\caption{A case study on the reasoning trace of RoE prompts highlights their key characteristic of explicitly incorporating structured legal reasoning aligned with HSE compliance requirements. }
\label{tab:case_study_3}

\begin{tabular}{m{1.3cm} m{12.2cm}}
\hline
\cellcolor{lightgray}{Types} & \cellcolor{lightgray}{Content} \\ \hline
\hline
Questions & This task involves evaluating an HSE legal or safety issue using multiple expert roles tailored to the case. Each expert provides an independent opinion using IRAC (Issue, Rule, Application, Conclusion) and step-by-step reasoning. A final judgment then synthesizes their insights into a justified, well-reasoned decision that balances regulatory compliance with practical safety. Based on the above instruction, please answer the following question: At a drilling site in Queensland, an incident occurred involving Jacob Kilby, a driller for Saxon Energy Services, who operated the Iron Roughneck in a manner that resulted in a fatal accident involving his colleague, Gareth Dodunski. Despite existing safety protocols, Saxon Energy Services faces scrutiny over their accountability for Jacob's actions, as well as questions about Carl Thomas, another operator, who did not update the safety management plan after a similar incident at another location. Considering these factors, what conclusion can be drawn regarding Saxon Energy's liability? (Evaluation without options)

\textbf{A}. Saxon Energy is not liable at all.

\textbf{B. Saxon Energy is liable due to Kilby's actions being within his employment scope.}

\textbf{C}. Saxon Energy's liability is negated by contractual clauses.

\textbf{D}. Saxon Energy is liable only if Kilby was unsupervised. 
\\ \hline

DeepSeek-R1

Reasoning Trace & Okay, I need to figure out Saxon Energy's liability for the fatal accident involving Jacob Kilby and Gareth Dodunski. Let me start by breaking down the problem.

{\color{orange}{First, the incident}}: Jacob operated the Iron Roughneck in a way that caused a fatal accident. There are existing safety protocols, but they're under scrutiny. Also, Carl Thomas didn't update the safety management plan after a similar incident elsewhere. So, Saxon might be liable either because of Jacob's actions....

{\color{orange}I remember in legal terms}, employers can be held vicariously liable for their employees' actions if those actions were within the scope of employment. But if the employee was acting negligently or against company protocols, does that absolve the employer? Maybe not entirely if the protocols were inadequate or not enforced properly...

{\color{orange}Under WHS laws in Australia, particularly the Work Health and Safety Act 2011,} employers have a primary duty of care. They need to ensure, so far as is reasonably practicable, the health and safety of workers. This includes maintaining and updating safety management systems, especially after incidents that highlight potential risks.

So, two angles: direct liability for not having proper protocols or training, and vicarious liability for Jacob's actions. Also, the failure to update the safety plan after a prior similar incident might show a pattern of negligence.

Using IRAC for each expert. Let's think of different expert roles. Maybe a Workplace Health and Safety Lawyer, a Safety Management Consultant, and a Vicarious Liability Legal Expert.

{\color{orange}Workplace Health and Safety Lawyer} would focus on the statutory duties under WHS laws. The issue is whether Saxon met their duty. The rule is the WHS Act 2011. Application: Saxon had prior incident, didn't update plan, leading to repeat incident. Conclusion: Breached duty, liable.

{\color{orange}Safety Management Consultant} would look at the systems in place. Issue: adequacy of safety management. Rule: Best practices require updating plans after incidents. Application: Failure to update after similar incident means system was deficient. Conclusion: Saxon's system failed, contributing to accident.

{\color{orange}Vicarious Liability Expert:} Issue is whether Saxon is liable for Jacob's actions. Rule: Employer liable if employee acts within scope. Application: Jacob was performing his job, even if negligently. Unless he was grossly negligent or acting outside duties, Saxon could be vicariously liable. But also, if Saxon's training/supervision was lacking, that's their direct liability.

Putting it all together, Saxon likely liable both for failing to update safety plans (systemic failure) and vicariously for Jacob's actions.
 \\ \hline

\hline
\end{tabular}

\end{table}
\begin{table}
\centering
\caption{A case study to validate the effectiveness of adversarial contextual augmentation. }
\label{tab:case_study_4}

\begin{tabular}{m{1.3cm} m{11.2cm}}
\hline
\cellcolor{lightgray}{Types} & \cellcolor{lightgray}{Content} \\ \hline
\hline
Original Question & In the context of the Code of Federal Regulations Title 29 Labor, an employer in California is charged with unfair labor practices. An Administrative Law Judge issues a decision recommending affirmative action by the the employer to address these practices. This decision is subsequently filed with the Board for approval. During the proceedings, the employer seeks to use the Alternative Dispute Resolution (ADR) Program to settle the dispute. What is the primary legal issue in this scenario?
\\ \hline

Augmented Question & During a recent compliance review at a manufacturing facility in California, it was discovered that several employees had filed complaints regarding unfair labor practices. An internal investigation led to a hearing, where an Administrative Law Judge recommended corrective measures to be undertaken by the employer. This recommendation was then forwarded to the Board for further consideration. In parallel, the employer expressed interest in resolving the matter through an internal mediation process, citing potential benefits of an expedited resolution. What is the primary legal issue in this scenario?\\ \hline
Options & 
\textbf{A. Whether the use of the Alternative Dispute Resolution (ADR) Program is mandatory for resolving unfair labor practices.}

\textbf{B}. Whether the decision of the Administrative Law Judge can be appealed by the employer.

\textbf{C}. Whether the Board must approve the Administrative Law Judge's decision before it is enforced.

\textbf{D}. Whether the employer's participation in the ADR Program waives any legal rights related to the case. 
\\ \hline

GPT-4o-mini on 

Original & A: Whether the use of the Alternative Dispute Resolution (ADR) Program is mandatory for resolving unfair labor practices. {\color{orange}(Correct)} \\ \hline

GPT-4o-mini on 

Augmented& C: Whether the Board must approve the Administrative Law Judge's decision before it is enforced.{\color{orange}(Incorrect)}\\ \hline

\hline
\end{tabular}

\end{table}


\end{document}